\documentclass{article}
\usepackage{textcomp}

 \usepackage[main, final]{neurips_2026}

\usepackage[utf8]{inputenc} 
\usepackage[T1]{fontenc}    
\usepackage{url}            
\usepackage{booktabs}       
\usepackage{amsfonts}       
\usepackage{nicefrac}       
\usepackage{microtype}      
\usepackage{xcolor}         

\usepackage{graphicx}
\usepackage{booktabs}

\usepackage[accsupp]{axessibility}  
\usepackage{wrapfig}
\usepackage{algorithm}
\usepackage{algpseudocode}

\usepackage{xspace}
\definecolor{cvprblue}{rgb}{0.21,0.49,0.74}
\usepackage[pagebackref,breaklinks,colorlinks,allcolors=cvprblue]{hyperref}
\usepackage{multirow}
\usepackage{amssymb} 
\usepackage{array}
\usepackage{tabularx} 
\usepackage{booktabs} 
\usepackage{amsmath}
\usepackage{subcaption}
\usepackage{graphicx}
\usepackage{makecell}
\usepackage{caption}
\usepackage{enumitem}
\usepackage{arydshln}

\usepackage{tcolorbox}

\newcommand{\ie}{\emph{i.e.}\xspace}
\newcommand{\eg}{\emph{e.g.}\xspace}

\newcommand{\sh}[1]{\textcolor{red}{{[#1]}}}

\title{Hide to See: Reasoning-prefix Masking for Visual-anchored Thinking in VLM Distillation}

\author{%
  Seonghoon Yu\textsuperscript{1}
  \quad
  Dongjun Nam\textsuperscript{3}
  \quad
  Byung-Kwan Lee\textsuperscript{2,$\dagger$}
  \quad
  Jeany Son\textsuperscript{3,$\dagger$}
  \\[0.5mm]
  \textsuperscript{1}KAIST
  \quad \quad
  \textsuperscript{2}NVIDIA
  \quad \quad
  \textsuperscript{3}POSTECH
  \\[0.5mm]
  {\small \texttt{seonghoon.yu@kaist.ac.kr}
  \quad
  \texttt{byungkwanl@nvidia.com}
  \quad
  \texttt{\{june6423,~jeany\}@postech.ac.kr}}
  \\[0.5mm]
  {\small Project Page:~\href{https://seonghoon-yu.github.io/Masking-KD-Page}{\texttt{https://seonghoon-yu.github.io/Masking-KD-Page}}}
}
\vspace{-0.1cm}

\begin{document}

\maketitle

\vspace{-3mm}
\begingroup
\renewcommand{\thefootnote}{\textdagger}
\makeatletter
\renewcommand{\@makefnmark}{\hbox{\normalfont\@thefnmark\hspace{0.3em}}}
\makeatother
\footnotetext{\vspace{-1mm}Corresponding authors.\vspace{-1mm}}
\endgroup
\vspace{-3mm}

\begin{abstract}
\vspace{-0.1cm}
Recent think-answer approaches in VLMs, such as Qwen3-VL-Thinking, boost reasoning performance by leveraging intermediate thinking steps before the final answer,
but their computational cost becomes substantial, especially for larger VLMs.
To distill such capabilities into compact think-answer VLMs, a primary objective is to improve the student's ability to utilize visual evidence throughout its reasoning trace, as long think-answer traces suffer from visual forgetting issues.
To this end, we introduce a novel think-answer distillation framework that encourages the student to anchor its thinking on visual information by masking the student's salient reasoning prefixes.
To compensate for such masked textual cues, the student is encouraged to rely more on visual evidence as an alternative source of information during distillation.
Our masking strategies include:
1) \textit{token-wise salient reasoning-prefix masking}, which masks high-influence reasoning prefixes selectively for each next-token prediction, and
2) \textit{self-paced masking budget scheduling}, which gradually increases the masking scale according to distillation difficulty, measured by the discrepancy between teacher--student distributions.
In the distillation phase, the student is guided by our salient reasoning-prefix mask, which blocks both future tokens and salient reasoning cues, in place of the standard causal mask used for auto-regressive language modeling.
Experimental results show that our approach outperforms recent open-source VLMs, VLM distillation, and self-distillation methods on multimodal reasoning benchmarks, while further analyzes confirm enhanced visual utilization along the student thinking process.

\end{abstract}
\vspace{-0.1cm}

\vspace{-0.2cm}

\section{Introduction}
\label{sec:intro}

\vspace{-0.1cm}

Recent think-answer approaches in vision-language models (VLMs) and large language models (LLMs), such as ChatGPT 5.4~\cite{openai_gpt54}, Gemini 3.1 Pro~\cite{gemini_41}, and DeepSeek-R1~\cite{deepseek-r1}, have achieved strong reasoning performance by explicitly generating reasoning before producing final answers.  
This paradigm has been widely adopted in subsequent VLMs~\cite{R1_onevision, rethinker, papo, recursive, r1_zero} to improve reasoning ability, particularly in complex problem-solving tasks such as mathematical reasoning~\cite{mathverse, mathvista} and scientific reasoning~\cite{mmmu_pro, science}.
However, the computational cost is particularly high for larger VLMs, limiting their deployment in resource-constrained scenarios such as on-device applications.

\begin{figure}[t]
    \centering

\includegraphics[width=\linewidth]{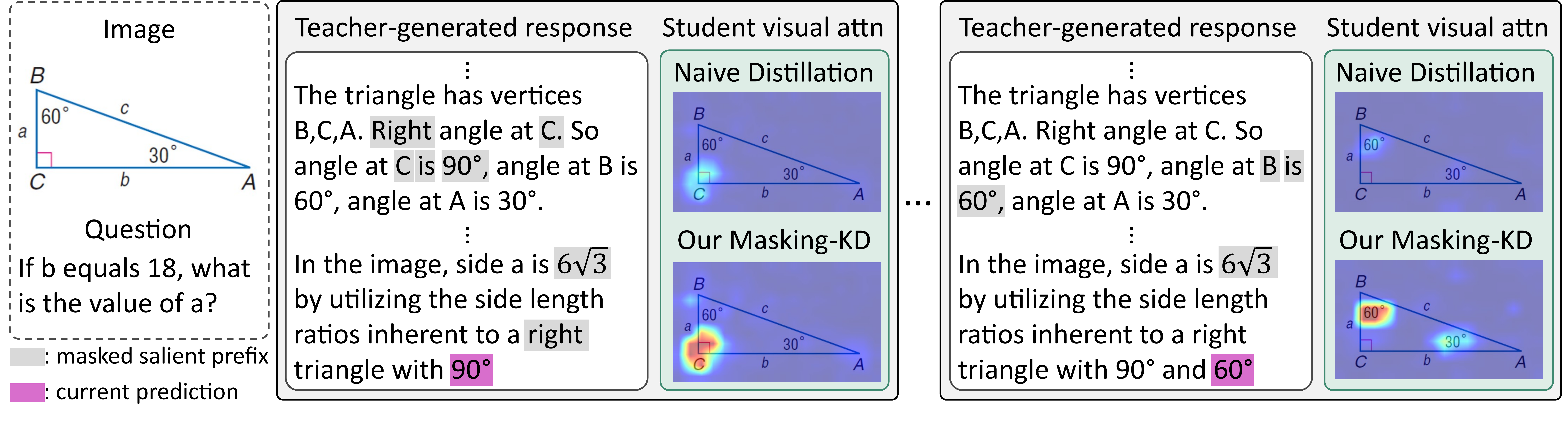}
    \vspace{-0.5cm}
    \caption{
    The illustration of our reasoning-prefix masking during VLM distillation.
    With full context of teacher's thinking trace (\ie, na\"ive distillation)\vspace{-0.2mm}, the student relies heavily on exposed textual prefixes to predict the \colorbox{purple!40}{current token}\vspace{-0.5mm}, resulting in weak visual attention.
    In contrast, with \colorbox{gray!20}{masked salient reasoning prefixes}\vspace{-0.4mm} (\ie, our Masking-KD), the student exploits more visual evidence to compensate for missing textual reasoning cues, improving its visual-anchored thinking.
    }
        \vspace{-0.3cm}
    \label{fig:intro}
\end{figure}

Knowledge distillation (KD) has emerged as a practical approach for reducing the computational overhead of large VLMs by transferring their capabilities to compact student models.
Since the distinctive capability of VLMs lies in connecting visual inputs with language outputs, effective distillation should preserve the student's reliance on visual evidence when producing language predictions.
Existing methods~\cite{llava-kd, compodistill, align_kd, switch_kd, align_kd_visual} mainly address this goal by transferring the teacher's visual knowledge through visual attention maps~\cite{compodistill}, vision token relations~\cite{llava-kd}, or vision projector alignment~\cite{align_kd_visual}.
While effective, these objectives primarily encourage the student to mimic the teacher's internal visual patterns, rather than guiding the student to anchor its own reasoning process in visual evidence.
This gap becomes particularly problematic for think-answer VLMs, where long think-answer trajectories expose rich reasoning cues that themselves provide enough information to predict subsequent tokens (analyzed in Sec.~\ref{subsec:effect_masking}), 
thereby diminishing the necessity to maintain sufficient visual reference throughout the thinking process, leading to visual forgetting~\cite{more_thought_less_acc, papo, pixel_reasoner}.

\begin{wrapfigure}{r}{0.35\textwidth}
    \centering
    \vspace{-0.4cm}
    \includegraphics[width=\linewidth]{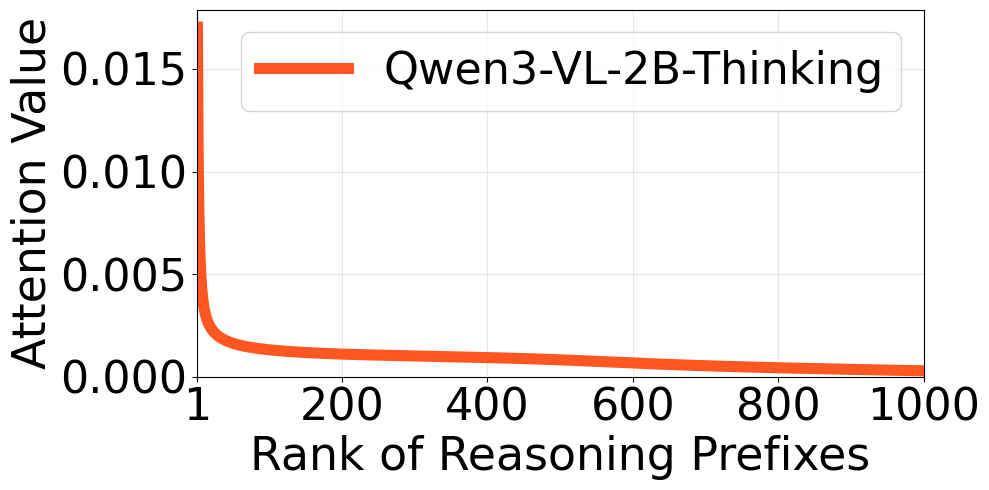}
    \vspace{-0.53cm}
    \caption{Reliance on salient cues}
    \vspace{-0.4cm}
    \label{fig:intro_warp}
\end{wrapfigure}
In particular, when distilling such long traces of think-answer VLMs, the student relies heavily on a small set of exposed textual cues, which receive disproportionately high attention values (Fig.~\ref{fig:intro_warp}), at every decoding step.
This suggests that only a few reasoning cues already provide sufficient information to follow the teacher's think-answer trace, which may reduce the student's need to learn from visual input.
This motivates our key question: 
\textit{If salient reasoning prefixes allow the student to imitate the teacher with less reliance on the image, can masking such prefixes encourage the student to exploit more visual evidence as an alternative to the masked salient textual cues?}
In this paper, we introduce Masking-KD, a novel distillation framework for think-answer models that enhances the student's ability to anchor its thinking in visual evidence by masking the student's prefixed reasoning cues, thereby encouraging it to rely more on visual cues to compensate for missing textual evidence (Fig.~\ref{fig:intro}).
Our salient reasoning-prefix mask is carefully constructed via:
1) \textbf{token-wise salient reasoning-prefix masking}, which identifies and masks the high-attention (\ie, salient) reasoning-context tokens selectively for each next-token prediction, since the most influential contextual cues vary across decoding steps, and masking them in a token-wise different manner;
and 2) \textbf{self-paced masking budget scheduling}, adaptively adjusts the masking scale for each next-token prediction according to its teacher-student KL divergence, assigning stronger masking to tokens that are easily imitated to amplify weak learning signals.
During distillation, the student is guided by our salient reasoning-prefix mask, which limits access to both future tokens and salient reasoning cues, instead of using the standard causal mask in auto-regressive language modeling.

In our experiments, the proposed framework outperforms recent open-source VLMs and VLM distillation methods on multimodal reasoning benchmarks. 
It also demonstrates effectiveness in self-distillation, where the student serves as its own teacher rather than a stronger teacher.
Furthermore, our analysis shows that masking salient reasoning prefixes during distillation improves the student's ability to derive its thinking from visual evidence.
Our contributions are summarized as follows:



\vspace{-0.1cm}
\begin{itemize}[leftmargin=10pt]
    \item We introduce Masking-KD, a novel think-answer distillation framework that enhances the student's ability to ground its thinking in visual evidence by masking the student's prefixed reasoning-context, encouraging the student to draw more on visual sources to compensate for missing textual cues.

    
    \item We propose \textit{token-wise salient reasoning-prefix masking} that identifies and masks high-influence prefixed reasoning cues selectively for each next-token prediction, reflecting that the most influential reasoning-context varies across every decoding step and masking them token-wise differently.
    


    \item We present \textit{self-paced masking budget scheduling} that adaptively adjusts the masking budget for each next-token prediction based on its teacher-student KL divergence, applying more aggressive masking on easily imitated tokens for the student so as to strengthen weak distillation signals.

    

    \item Extensive experiments validate the effectiveness of the proposed framework by surpassing recent open-source VLMs, VLM distillations, and self-distillations on multimodal reasoning benchmarks.
    Further analysis shows that it improves the student’s visual-anchored thinking from diverse aspects.
    

\end{itemize}

\vspace{-0.1cm}
\section{Think-Answer Reasoning Distillation Framework}
\label{sec:method}
\vspace{-0.1cm}

In this section, we present Masking-KD, a simple yet effective think-answer distillation that masks the accumulated salient reasoning prefixes of the student to 
guide its thinking process based on visual evidence. 
Unlike existing VLM distillation methods~\cite{compodistill, llava-kd, align_kd}, which directly transfer the teacher's visual knowledge through visual attention maps or visual token relations, Masking-KD encourages the student to develop its own visual-anchored thinking via salient reasoning-prefix masking.
By mimicking teacher-like distributions under missing textual cues, the student learns to rely more on visual evidence, providing valuable learning signals beyond conventional teacher–student alignment.






We begin with an overview of Masking-KD (Sec.~\ref{subsec:overall_framework}), explaining how the student is distilled under masked reasoning cues.
We then describe how the salient reasoning-prefix mask is constructed: 1) \textit{token-wise salient reasoning-prefix masking} (Sec.~\ref{subsec:token-wise}), determining which reasoning prefixes are masked, and 2) \textit{self-paced masking budget scheduling} (Sec.~\ref{subsec:self-paced}), deciding how many reasoning prefixes are masked.
Both are defined adaptively for each next-token prediction and are implemented via an attention mask under the standard causal masking~\cite{attention_is_all} used for auto-regressive language modeling, supporting token-wise different masking in a single forward pass.

\begin{figure}[t]
    \centering

\includegraphics[width=0.95\linewidth]{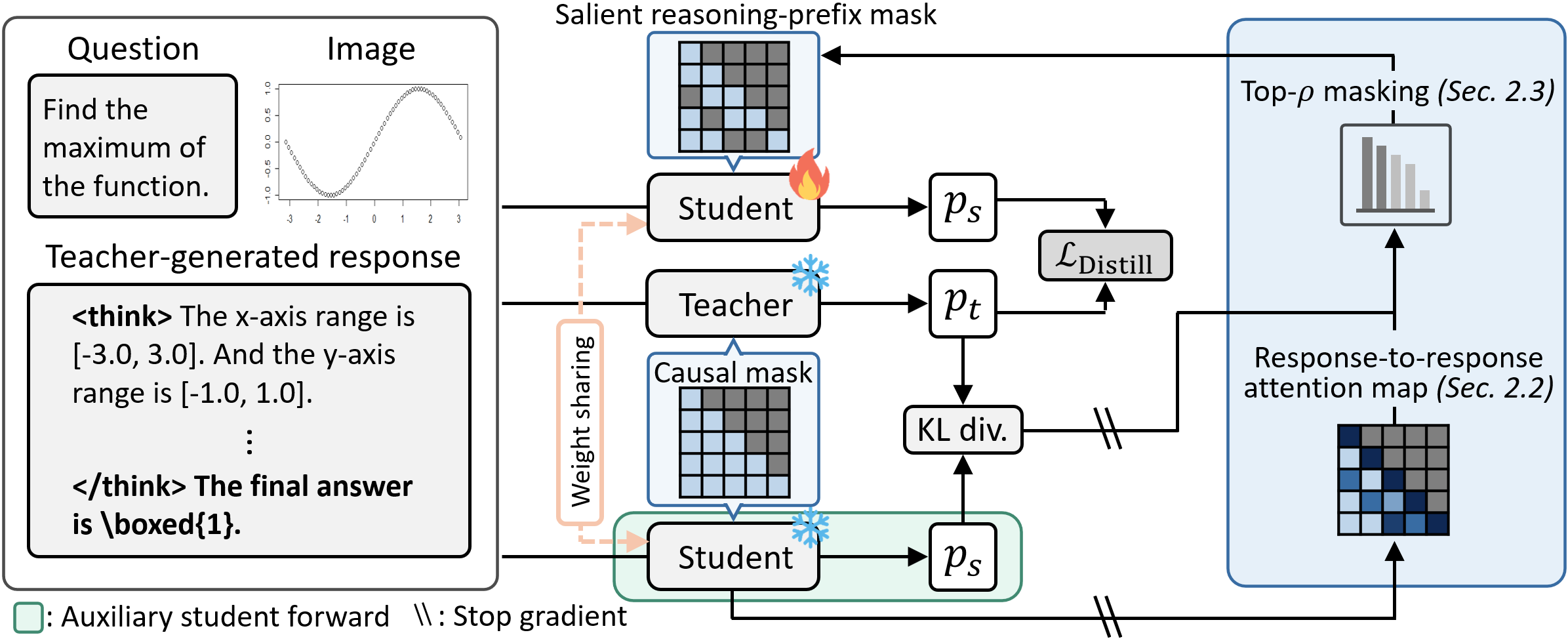}
    \caption{The illustration of Masking-KD. During distillation, the student is guided by our salient reasoning-prefix mask that blocks access to both future tokens and salient reasoning prefixes, whereas the teacher operates under the causal mask. 
    This salient reasoning-prefix mask is derived from two quantities extracted from auxiliary student forward under the causal mask: 1) a response-to-response attention map for identifying salient reasoning-prefix (Sec.~\ref{subsec:token-wise}), and 2) token-wise reverse KL divergence for adaptively deciding the masking strength for each token (Sec.~\ref{subsec:self-paced}).
    }
    \vspace{-0.3cm}
    \label{fig:method}
\end{figure}

\vspace{-0.1cm}
\subsection{Overview of Masking-KD}
\label{subsec:overall_framework}
\vspace{-0.1cm}

Our knowledge distillation framework (Fig.~\ref{fig:method}) is built upon reverse KL divergence~\cite{gkd} to align the student's predictive distribution $p_s$ over the vocabulary $\mathcal{V}$ with the teacher's one $p_t$ along the distilled token sequence $\mathbf{y}=\{y_1,\dots,y_N\}$ of length $N$, given the input image $\mathbf{x}_v$ and question $\mathbf{x}_q$.
To impose masked reasoning prefixes on the student, we modify the causal mask used in the auto-regressive language modeling so that, at each decoding step, the student is restricted from access to both future tokens and salient reasoning-context tokens, as follows:
\begin{equation}
\label{eq:reverse_kl}
\mathcal{L}_{\text{Distill}}
= \frac{1}{N}\sum_{n=1}^{N}\sum_{y\in\mathcal{V}} p_s({y} \mid \mathbf{x}_v,\mathbf{x}_q,\mathbf{y}_{<n},\tilde{\mathbf{M}})\log\frac{p_s({y} \mid \mathbf{x}_v,\mathbf{x}_q,\mathbf{y}_{<n},\tilde{\mathbf{M}})}{p_t({y} \mid \mathbf{x}_v,\mathbf{x}_q,\mathbf{y}_{<n},\mathbf{M})}.
\end{equation}
Here, $p_{s}$ and $p_t$ are scaled by the distillation temperature $\tau$ and $\mathbf{y}_{<n}$ indicates the prefixed reasoning-context tokens $\{y_1,\dots,y_{n-1}\}$ up to step $n$.
$\mathbf{M}$ is the standard causal mask used for the teacher, while $\tilde{\mathbf{M}}$ indicates the salient reasoning-prefix mask for the student, which extends the causal mask $\mathbf{M}$ by additionally masking salient reasoning-prefix tokens, enforcing the student to infer each subsequent token with missing prefixes during distillation. 

\vspace{-0.1cm}
\paragraph{Auxiliary Student Forward.}
\label{sec:auxiliary}
To construct the salient reasoning-prefix mask $\tilde{\mathbf{M}}$, we perform an auxiliary forward pass of the student $\theta_s$ under the vanilla causal mask $\mathbf{M}$, \ie, $\theta_s(\mathbf{x}_v,\mathbf{x}_q,\mathbf{y},\mathbf{m})$, where we extract two quantities for mask construction:
(1) \textbf{response-to-response attention map} ${\mathbf{A}}^{\text{resp}}$, used to identify which prefixes to mask (Sec.~\ref{subsec:token-wise}), and (2) \textbf{token-wise reverse KL divergence} $\mathbf{r}=\{r_n\}_{n=1}^N,$, used to determine how many prefixes to mask (Sec.~\ref{subsec:self-paced}). 
Specifically, we compute these two quantities as follows.

\textbf{(1) response-to-response attention map} ${\mathbf{A}}^{\text{resp}}$:
 To obtain this, we first average attention maps across all $H$ transformer layers
and then restrict the averaged attention map to the response-token block:
\begin{equation}
\label{eq:response_attn}
\mathbf{A}^{\text{resp}} \leftarrow \mathbf{A}\!\restriction_{\mathcal{I^{\text{resp}}}\times\mathcal{I^{\text{resp}}}}\in\mathbb{R}^{N\times N},\quad 
\text{where~}\mathbf{A} = \frac{1}{H}\sum_{h=1}^H\mathrm{Attn}^h(\hat{\mathbf{x}}^{h-1}_v,\hat{\mathbf{x}}^{h-1}_q,\hat{\mathbf{y}}^{h-1},\mathbf{M}),
\end{equation}
$\text{Attn}^h(\cdot)$ denotes the attention operation at the $h$-th transformer layer, $\restriction_{\mathbf{I}\times\mathbf{J}}$ indicates the submatrix restriction to the rows indexed by $\mathbf{I}$ and columns indexed by $\mathbf{J}$, and $\mathcal{I}^{\text{resp}}$ denotes the index set of response-token positions.
In particular, $\hat{\mathbf{x}}^{h-1}_v$, $\hat{\mathbf{x}}^{h-1}_q$, and $\hat{\mathbf{y}}^{h-1}$ denote the visual, question, and response representations at the input of the $h$-th transformer layer, respectively. 
Under the causal mask $\mathbf{M}$, the first $n-1$ entries of $n$-th row from $\mathbf{A}^{\text{resp}}$ represent the attention values over prefixed textual tokens $\mathbf{y}_{<n}=\{y_1, \dots, y_{n-1}\}$ used to predict $y_n$.
We leverage these attention weights to identify the salient reasoning-prefix tokens in Sec.~\ref{subsec:token-wise}.

\textbf{(2) token-wise reverse KL divergence} $\mathbf{r}$:
Under the standard causal mask $\mathbf{M}$, we calculate the reverse KL divergence between the student distribution $p_s$ and the teacher distribution $p_t$ over the vocabulary $\mathcal{V}$ at every response position, where $p_t$ is reused from Eq.~\eqref{eq:reverse_kl}.
Formally, this is given by:
\begin{equation}
\label{eq:r}
\mathbf{r}=\{r_n\}_{n=1}^N,\quad\text{where~} r_n=\sum_{y\in\mathcal{V}}p_s({y} \mid \mathbf{x}_v,\mathbf{x}_q,\mathbf{y}_{<n},{\mathbf{M}}) \log \frac{
p_s({y} \mid \mathbf{x}_v,\mathbf{x}_q,\mathbf{y}_{<n},{\mathbf{M}})}{p_t({y} \mid \mathbf{x}_v,\mathbf{x}_q,\mathbf{y}_{<n},\mathbf{M})}.
\end{equation}
Here, $\mathbf{r}$ captures the distributional discrepancy between the teacher and the student at each token position over $N$ response tokens, reflecting the distillation difficulty of every token in the distilled response.
We use this quantity to determine the masking strength for each token in Sec.~\ref{subsec:self-paced}.

\vspace{-0.1cm}
\subsection{Token-wise Salient reasoning-prefix Masking}
\vspace{-0.1cm}
\label{subsec:token-wise}
Throughout the distilled thought-answer trace, the student tends to rely heavily on a small subset of salient reasoning cues with disproportionately high attention, as illustrated in Fig.~\ref{fig:intro_warp}.
By blocking access to such salient prefixes for the student, we promote the use of visual information to compensate for the masked salient reasoning cues.
To this end, we propose \textit{token-wise salient reasoning-prefix masking}, which selectively masks salient prefixes for each next-token prediction to reflect that the most influential contextual tokens differ across decoding steps.


\vspace{-0.2cm}
\paragraph{Construct Salient Reasoning-prefix Mask.}
To create a salient reasoning-prefix mask $\tilde{\mathbf{M}}$ used for distillation in Eq.~\eqref{eq:reverse_kl}, we utilize the response-to-response attention map $\mathbf{A}^{\text{resp}}$ extracted from an auxiliary student forward (Sec.~\ref{sec:auxiliary}).
At each decoding step $n$, the row of this map $\mathbf{A}^{\text{resp}}_{n,<n}$ captures how strongly the preceding textual tokens $\mathbf{y}_{<n}=\{y_1,\dots,y_{n-1}\}$ contribute to predict the current token $y_n$.
Based on this prefix attention, we collect salient tokens using a nucleus top-$p$~\cite{top_p} style rule, which we refer to as top-$\rho$ masking, greedily selecting the highest-attended prefixes until their cumulative attention ratio reaches $\rho_n$.
The collected prefixes form a salient prefix set $\mathcal{S}_n$ such that:
\begin{equation}
    \label{eq:cumulative}
    {\sum_{j\in\mathcal{S}_n}\mathbf{\bar{A}}^{\text{resp}}_{n,j}} \geq \rho_n, \quad \text{where}~\bar{\mathbf{A}}^{\text{resp}}_{n,j}=\frac{\mathbf{A}^{\text{resp}}_{n,j}}{\sum_{k=1}^{n-1}\mathbf{A}^{\text{resp}}_{n,k}}.
\end{equation}
Here, $\mathbf{\bar{A}}^{\text{resp}}_{n,j}$ denotes the attention score assigned to the $j$-th prefix token, normalized over all $n-1$ prefixes.
The threshold $\rho_n$ is a self-paced cumulative ratio for step $n$ (introduced in Sec.~\ref{subsec:self-paced}).
Full details are provided in Appendix~\ref{app:subsec:top-rho}.
We then construct the salient reasoning-prefix mask $\tilde{\mathbf{M}}\in\{-\infty,0\}^{N\times N}$ by extending the standard causal mask $\mathbf{M}$ to additionally suppress the salient prefix position in $\mathcal{S}_n$.
Specifically, each entry of the salient reasoning-prefix mask $\tilde{\mathbf{M}}$ is given by:
\begin{equation}
    \label{eq:mask}
    \tilde{\mathbf{M}}_{n,j} = \left\{
    \begin{array}{lll}
        -\infty, & \text{if } j > n & \text{(Causal masking)} \\
        -\infty, & \text{if } j \in \mathcal{S}_n  & \text{(Salient prefix masking)} \\
        0, & \text{otherwise} & 
    \end{array}
    \right., \quad \forall n \in \{1, \dots, N\}.
\end{equation}
The resulting $\tilde{\mathbf{M}}$ is added to the attention logits as an attention mask, so that when predicting the current token $y_n$, the student is prevented from attending to future tokens and salient reasoning prefixes.
For clarity, we omit visual and question token positions from both this formula and Fig~\ref{fig:method}.

\vspace{-0.1cm}
\subsection{Self-Paced Masking Budget Scheduling}
\vspace{-0.15cm}
\label{subsec:self-paced}
As reasoning prefixes gradually accumulate, the student can imitate the teacher’s subsequent tokens more easily, since prefixes themselves provide increasingly sufficient information (analyzed in Sec.~\ref{subsec:effect_masking}).
This suggests that applying a uniform masking budget to every response token is suboptimal.
To address this, we introduce \textit{self-paced masking budget scheduling}, which adaptively allocates masking budget (\ie, the amount of salient reasoning-prefix to mask) according to the distillation difficulty of each token.
As a result, easier tokens (\ie, those with lower distillation loss) receive stronger masking to recover weakened distillation signals, while already harder tokens (\eg, those with higher distillation loss) have greater access to reasoning prefixes.

\vspace{-0.1cm}
\paragraph{Self-Paced Cumulative Ratio Threshold.} 
In our framework, the masking amount is controlled by a self-paced cumulative ratio threshold $\rho_n$ in Eq.~\eqref{eq:cumulative}. 
To determine this, we utilize token-wise reverse KL divergence $\mathbf{r}=\{r_n\}_{n=1}^N$ obtained from the auxiliary student forward in Eq.~\eqref{eq:r}.
Here, $\mathbf{r}$ captures the distillation difficulty along the distilled trace of length $N$.
From $\mathbf{r}$, we decide $\rho_n$, as follows:
\begin{equation}
\rho_n=\rho_{\min} + (\rho_{\max}-\rho_{\min}) \cdot \sigma\!\left(\tilde{r}_n-\mu_{\tilde{r}}\right),\quad \text{where} ~\tilde{r}_n = -\log(r_n+\epsilon), \quad \mu_{\tilde{r}}= \frac{1}{N}\sum_{i=1}^N \tilde{r}_i,
\end{equation}
$\rho_\text{min}$ and $\rho_{\text{max}}$ denote pre-defined lower and upper bounds of $\rho_n\in[\rho_{\text{min}},\rho_{\text{max}}]$,  which control the overall masking amount.
$\sigma(\cdot)$ is the sigmoid function that maps the score to $[0,1]$.
We transform each $r_n$ into a log-scaled score $\tilde{r}_n=-\log(r_n+\epsilon)$ to compress the dynamic range of $\{r_n\}_{n=1}^N$, which stabilizes the resulting threshold $\rho_n$.
The negative sign in $\tilde{r}_n=-\log(r_n+\epsilon)$ reverses the ordering of the scores, so that tokens with smaller reverse KL values receive larger $\rho_n$.
We subtract the mean score $\mu_{\tilde{r}}$ before applying $\sigma$ to center $\{\rho_n\}_{n=1}^N$ around the average of $\mathbf{r}$.
As a result, tokens with average difficulty are mapped near the midpoint of $[\rho_{\text{min}},\rho_{\text{max}}]$, while relatively harder and easier tokens are pushed toward $\rho_{\text{min}}$ and $\rho_{\text{max}}$, respectively.




\vspace{-0.2cm}
\section{Experiments}
\label{sec:experiments}
\vspace{-0.15cm}

\subsection{Experimental Setup}
\vspace{-0.15cm}
\label{subsec:experiments_setup}

\paragraph{Dataset and Metric.}
For VLM distillation, we construct the distilled data from ViRK39K~\cite{rethinker} dataset by extracting the teacher's think-answer traces using greedy decoding with a maximum length of 4096 tokens, using the instructions in Appendix~\ref{app:instruction}. 
For self-distillation, the student uses its own pre-extracted think-answer traces.  
In all cases, we keep only correct responses, yielding 19k, 15k, and 10k samples for the 8B, 4B, and 2B Qwen3-VL-Thinking~\cite{qwen3_vl}, respectively.
For evaluation, we report pass@1 results with a maximum generation of 4096 tokens on:
1) math and geometric reasoning: Geometry-3K~\cite{geo3k}, MathVista~\cite{mathvista}, We-Math~\cite{wemath}, MMK12~\cite{mmeureka}, MathVerse~\cite{mathverse};
2) logical reasoning: LogicVista~\cite{logicvista};
and 3) multi-discipline multimodal reasoning: MMMU-Pro~\cite{mmmu_pro}.

\begin{table}[t]
    \centering
    \scriptsize
    \renewcommand{\arraystretch}{1.2}
    \setlength{\tabcolsep}{4.1pt}   
    \caption{\textbf{Comparison with open-source VLMs.} 
    Our Masking-KD employs Qwen3-VL-Thinking models.
    $\dagger$ denotes the self-distilled 8B model using its own self-teacher, and $\ddagger$ indicates the distilled student from the 8B teacher.
    We evaluate all compared VLMs using greedy decoding with a maximum length of 4096 for direct comparison.
    Results on other VLM models are provided in Appendix~\ref{app:subsec:other_vlms}.
    }
    \vspace{0.1cm}
    \scalebox{1.055}{
    \begin{tabular}{m{3.3cm}|*{7}{>{\centering\arraybackslash}m{0.94cm}}|>{\centering\arraybackslash}m{0.7cm}}
    \hline
     Method     &  Geo3k & MathVista & We-Math & MMK12 & MathVerse & LogicVista & MMMU$^\text{Pro}$ & Avg. \\ \cline{1-9}
    \multicolumn{9}{l}{\quad \textit{\textbf{$\sim$8B Models}}} \\ \cdashline{1-9}[0.2pt/1pt]
    
    {Qwen3-VL-8B-Thinking}  & 54.58&	65.20&	66.15&	42.55&	63.81&	43.40&	39.83&	53.65 \\ 
    Ovis2-8B~\cite{ovis} & 42.43&	68.20&	64.66&	48.15&	61.19&	43.62&	16.18&	49.20 \\
    InternVL3.5-8B~\cite{internvl3_5}& 44.59&	68.50&	56.61&	44.95	&53.26&	37.81&	38.50	&49.17 \\
    InternVL3-8B~\cite{internvl3} & 38.44	&53.20&	51.32&	39.80&	54.82&	47.43&	36.01&	45.86 \\
    MiMo-VL-8B~\cite{mimo}& 62.23	&69.50	&42.41&	44.10&	18.12&	46.76&	34.57&	45.38\\
    Qwen2.5-VL-7B~\cite{qwen2_5} & 40.43&	67.50	&48.74&	43.90&	38.67&	46.31	&31.56&	45.30\\
        \cdashline{1-9}[0.2pt/1pt]
    {\tiny{\textit{Self-distill}}} Masking-KD-8B$^{\dagger}$ (ours)  & \textbf{58.24}	&\textbf{67.10}	&\textbf{71.72}&	\textbf{49.95}	&\textbf{67.84}	&\textbf{48.10}&	\textbf{43.47}	&\textbf{58.06}\\
    \hline
    \multicolumn{9}{l}{~~~~\textit{\textbf{$\sim$4B Models}}} \\ \cdashline{1-9}[0.2pt/1pt]
    Qwen3-VL-4B-Thinking  & 43.93&	62.60&	49.37&	31.55&	49.86&	39.37&	32.08 & 44.11 \\
    Ovis2-4B~\cite{ovis} & 37.77	&61.10	&60.29&	39.10&	58.03&	39.60&	12.60&	44.07\\
    InternVL3.5-4B ~\cite{internvl3_5} &41.93&	52.10&	45.46&	25.80	&43.21	&27.29&	29.77&	37.94\\
    Qwen2.5-VL-3B~\cite{qwen2_5} & 26.29	&55.90&	49.66	&39.85&	40.69&	38.03&	27.75&	39.74\\
    \cdashline{1-9}[0.2pt/1pt]
    Masking-KD-4B$^{\ddagger}$ (ours)  & \textbf{52.58}	&\textbf{66.50}&	\textbf{71.03}&	\textbf{51.00}&	\textbf{62.66} & \textbf{52.35} & \textbf{40.52} & \textbf{56.66} \\
        \hline
    \multicolumn{9}{l}{~~~~\textit{\textbf{$\sim$2B Models}}} \\ \cdashline{1-9}[0.2pt/1pt]
    Qwen3-VL-2B-Thinking &  26.29&	43.10&	25.17&	13.00&	28.21&	18.57&	14.51&	24.12   \\ 
        Ovis2-2B~\cite{ovis}&31.11&	54.70&	51.95&	32.45&	50.32	&31.77&	10.23	&37.50 \\
    InternVL3.5-2B~\cite{internvl3_5} & 29.95	&40.60	&24.31&	14.70&	27.94&	16.11&	12.54&	23.74\\
        InternVL3-2B~\cite{internvl3} & 33.78&	46.70&	47.93	&37.00&	40.37	&33.33&	22.95&	37.44\\
    \cdashline{1-9}[0.2pt/1pt]
    {Masking-KD}-2B$^{\ddagger}$ (ours)  &  \textbf{40.93}&	\textbf{59.20}&	\textbf{63.79}&	\textbf{37.20}&	\textbf{57.89}&	\textbf{41.61}&	\textbf{30.75}&	\textbf{47.34} \\

    \hline
    \end{tabular}
    }
    \vspace{-0.4cm}
    \label{tab:open-source}
\end{table}

\begin{table}[t]
    \centering
    \scriptsize
    \renewcommand{\arraystretch}{1.2}
    \setlength{\tabcolsep}{5pt}   
    \vspace{-0.1cm}
    \caption{\textbf{Result on self-distillation.} The base model is distilled using its own predictions under each method.
    The details on self-distillation are elaborated in Appendix~\ref{app:subsec:self_distill}.
    }
    \vspace{0.1cm}
    \scalebox{1.055}{
    \begin{tabular}{l|ccccccc|c}
    \cline{1-9}
    Method     &  Geo3k & MathVista & We-Math & MMK12 & MathVerse & LogicVista & MMMU$^{\text{Pro}}$ & Avg. \\ \cline{1-9}
    Base \tiny{\textit{{Qwen3-VL-8B-Thinking}}} & \textcolor{gray}{54.58} & \textcolor{gray}{65.20} & \textcolor{gray}{66.15} & \textcolor{gray}{42.55} & \textcolor{gray}{63.81} & \textcolor{gray}{43.40} & \textcolor{gray}{39.83} & \textcolor{gray}{53.65} \\ \cdashline{1-9}[0.2pt/1pt]
    ~~w/ OPSD~\cite{opsd} &56.42&	66.20&	67.37&	44.62&	64.67&	45.29&	41.33&	55.13\\
    ~~w/ Masking-KD (ours)  & \textbf{58.24}	&\textbf{67.10}	&\textbf{71.72}&	\textbf{49.95}	&\textbf{67.84}	&\textbf{48.10}&	\textbf{43.47}	&\textbf{58.06}\\
        \hline
    Base \tiny{\textit{{Qwen3-VL-4B-Thinking}}}  & \textcolor{gray}{43.93} & \textcolor{gray}{62.60} & \textcolor{gray}{49.37} & \textcolor{gray}{31.55} & \textcolor{gray}{49.86} & \textcolor{gray}{39.37} & \textcolor{gray}{32.08} & \textcolor{gray}{44.11} \\ \cdashline{1-9}[0.2pt/1pt]
    ~~w/ OPSD~\cite{opsd} & 48.75&	60.80&	58.74&	35.10&	56.51&	39.82&	32.83&	47.51 \\
    ~~w/ Masking-KD (ours)  &\textbf{52.25}&	\textbf{66.10}&	\textbf{68.79}&	\textbf{50.85}&	\textbf{64.08}&	\textbf{50.78}&	\textbf{39.25}&	\textbf{56.01}\\
    \hline
        Base \tiny{\textit{{Qwen3-VL-2B-Thinking}}}&  \textcolor{gray}{26.29} & \textcolor{gray}{43.10} & \textcolor{gray}{25.17} & \textcolor{gray}{13.00} & \textcolor{gray}{28.21} & \textcolor{gray}{18.57} & \textcolor{gray}{14.51} & \textcolor{gray}{24.12} \\ \cdashline{1-9}[0.2pt/1pt]
    ~~w/ OPSD~\cite{opsd} & 26.46 & 44.20  & 26.09  & 14.00 & 31.79 & 19.24 & 15.84 & 25.37\\
    ~~w/ Masking-KD (ours)  & \textbf{33.61} & \textbf{52.00} & \textbf{40.40} & \textbf{19.25} & \textbf{40.09} & \textbf{25.28}  & \textbf{20.00} & \textbf{32.95} \\
        \hline
        
    \cline{1-9}

    \end{tabular}
    }
    \vspace{-0.4cm}
    \label{tab:self_distillation}
\end{table}

\vspace{-0.15cm}
\paragraph{Implementation Details.}
We build our framework on Qwen3-VL-Thinking~\cite{qwen3_vl}.
The student is trained for 2 epochs using a learning rate of $1\times10^{-6}$, with a batch size of 1 and gradient accumulation over 512 steps.
The auxiliary student forward (Sec.~\ref{sec:auxiliary}) uses a weight-shared student rather than a separately initialized one.
The self-paced cumulative ratio $\rho_n$ (Sec.~\ref{subsec:self-paced}) is bounded by $\rho_{\text{min}}=0.3$ and $\rho_{\text{max}}=0.5$.
To stabilize training and prevent loss explosion, we exclude the prefix token directly preceding the current token from masking.
We set the distillation temperature $\tau=2$ in the reverse KL divergence of Eq.~\eqref{eq:reverse_kl}.
All experiments are conducted on NVIDIA A100 80 GB GPUs: two GPUs for the 2B and 4B students, and four GPUs for the 8B student in the self-distillation.

\vspace{-0.2cm}
\subsection{Main Results}
\vspace{-0.15cm}

\paragraph{Comparison with Open-source VLMs .}
We compare Pass@1 results of our Masking KD with open-source VLM models~\cite{ovis, internvl3, internvl3_5, mimo, qwen2_5} on multimodal reasoning benchmarks in Tab.~\ref{tab:open-source}.
Masking-KD-8B is obtained by self-distilling Qwen3-VL-8B-Thinking using its own predictions,
while Masking-KD-4B and -2B are distilled from the Qwen3-VL-8B-Thinking teacher.
Because all compared VLMs report their results with different generation lengths, we re-evaluate them using greedy decoding with a maximum length of 4096 tokens to ensure a direct comparison. 
Our Masking-KD achieves state-of-the-art performance across all model sizes.
Notably, our compact 2B model outperforms the undistilled 4B model, and our 4B model surpasses the undistilled 8B model, indicating the effectiveness of ours.
Results on other VLM models are provided in Appendix~\ref{app:subsec:other_vlms}.




\vspace{-0.15cm}
\paragraph{Results on Self-Distillation.}
To validate the effectiveness of our masking approach in self-distillation settings, Tab.~\ref{tab:self_distillation} reports Pass@1 performance of our methods compared with another self-distillation method~\cite{opsd}.
In this experiment, each model is distilled using its own pre-extracted think-answer trajectories.
While OPSD~\cite{opsd} uses the student's thought-answer trace as input to construct a self-teacher signal, our method instead performs self-distillation with masked reasoning prefixes, forcing the model to recover its own predictions without relying on salient thinking cues.
The superior performance of our methods over OPSD confirms the effectiveness of salient prefix masking in self-distillation.
Further details on the self-distillation are provided in Appendix~\ref{app:subsec:self_distill}.



\vspace{-0.15cm}
\paragraph{Comparison with other VLM Distillations.}
To demonstrate the superiority of our Masking-KD, we compare its Pass@1 results with recent VLM distillation methods~\cite{llava-kd, compodistill, align_kd} on multimodal reasoning benchmarks.
Tab.~\ref{tab:main_vlm_distill_4b} and Tab.~\ref{tab:main_vlm_distill_2b} present results for the 8B teacher -- 4B student and 8B teacher -- 2B student configurations, respectively.
The na\"ive response distillation refers to the standard response-level distillation using reverse KL divergence.
Since all compared methods are designed for distilling instruction-following abilities from scratch, we reproduce them in our think-answer distillation settings.
In addition, these methods improve the student's visual perception ability by transferring the teacher's visual patterns, such as visual attention maps in CompoDistill~\cite{compodistill}, vision token relations in LLaVA-KD~\cite{llava-kd}, and instruction-aware visual focus in Align-TI~\cite{align_kd}.
Despite not explicitly distilling such visual patterns, Masking-KD achieves the best results, showing that reasoning-prefix masking is more effective in VLM reasoning distillation.
Results obtained by combining our method with other VLM distillation approaches are reported in Appendix~\ref{app:subsec:pnp_kd}.

\vspace{-0.15cm}
\paragraph{Extension to Supervised Fine-Tuning.}
Since salient reasoning-prefix masking does not inherently depend on the distillation scenario, we further apply it to standard supervised fine-tuning (SFT), which is useful when only teacher-generated responses are available (\eg, closed-source or API-only teachers).
In this setting, the 2B student is trained on the 8B teacher-generated responses with one-hot labels, and the token-wise reverse KL divergence in self-paced masking budget scheduling is replaced with the per-token cross-entropy loss.
As shown in Tab.~\ref{tab:sft}, our masking consistently improves SFT across all benchmarks, demonstrating its generality beyond distillation.

\vspace{-0.15cm}
\paragraph{Robustness to Perturbed Reasoning Prefixes.}
To verify that salient reasoning-prefix masking reduces the student's dependence on exposed reasoning prefixes, we intentionally perturb the reasoning prefixes at inference and compare the resulting performance degradation with na\"ive distillation.
Specifically, on 400 problems, we perturb the student's reasoning prefixes at every decoding step in two ways: replacing numerical values with incorrect ones (\textit{corrupted prefixes}; Tab.~\ref{tab:robustness_a}) and randomly removing prefix tokens (\textit{dropped prefixes}; Tab.~\ref{tab:robustness_b}).
As shown in Tab.~\ref{tab:robustness}, Masking-KD shows much smaller degradation than na\"ive distillation under both perturbations, indicating its reduced dependence on reasoning prefixes.

\begin{table}[t]
    \centering
    \scriptsize
    \renewcommand{\arraystretch}{1.2}
    \setlength{\tabcolsep}{4.1pt}   
        \caption{\textbf{Comparison with other VLM distillations (8B teacher -- 4B student).}
        We employ Qwen3-VL-Thinking for both teacher and student models.
    ${\dagger}$ denotes the methods proposed for distilling instruction-following ability; for this experiment, we reproduce them in our reasoning distillation setting.
    The results when combining our approach with them are reported in Appendix~\ref{app:subsec:pnp_kd}.}
        \vspace{0.1cm}
    \scalebox{1.055}{
    \begin{tabular}{m{3.2cm}|*{7}{>{\centering\arraybackslash}m{0.94cm}}|>{\centering\arraybackslash}m{0.7cm}}
    \cline{1-9}
    Method     &  Geo3k & MathVista & We-Math & MMK12 & MathVerse & LogicVista & MMMU$^{\text{Pro}}$ & Avg. \\ \cline{1-9}
    Teacher \tiny{\textit{Qwen3-VL-8B-Thinking}}  & 54.58&	65.20&	66.15&	42.55&	63.81&	43.40&	39.83&	53.65 \\
    Student \tiny{\textit{Qwen3-VL-4B-Thinking}} & 43.93&	62.60&	49.37&	31.55&	49.86&	39.37&	32.08 & 44.11\\
    
    \cline{1-9}
    Na\"ive Response Distillation & 47.42&	62.80&	60.00&	37.85&	56.97&	42.95&	34.22&	48.89\\ 
    \cdashline{1-9}[0.2pt/1pt]
    LLaVA-KD$^{\dagger}$~\cite{llava-kd} & 49.75&	63.80&	62.59	&39.45	&59.27	&43.40&	34.28	&50.36	 \\
    CompoDistill$^{\dagger}$~\cite{compodistill} & 50.92&	63.40&	64.54&	39.30	&60.18&	45.19	&35.49&	51.29 \\
    Align-TI$^{\dagger}$~\cite{align_kd} & 50.58  & 62.80	& 62.75	&39.15	& 58.30	&	42.95& 35.38	& 50.27 \\
    \cdashline{1-9}[0.2pt/1pt]
    Masking-KD (ours)  & \textbf{52.58}	&\textbf{66.50}&	\textbf{71.03}&	\textbf{51.00}&	\textbf{62.66} & \textbf{52.35} & \textbf{40.52} & \textbf{56.66} \\

    \hline
    \end{tabular}
    }
    \label{tab:main_vlm_distill_4b}
    \vspace{-0.43cm}
\end{table}

\begin{table}[t]
    \centering
    \scriptsize
    \renewcommand{\arraystretch}{1.2}
    \setlength{\tabcolsep}{4.1pt}   
        \caption{\textbf{Comparison with other VLM distillations (8B teacher -- 2B student).}
    }
    \vspace{0.1cm}
    \scalebox{1.055}{
    \begin{tabular}{m{3.2cm}|*{7}{>{\centering\arraybackslash}m{0.94cm}}|>{\centering\arraybackslash}m{0.7cm}}
    \cline{1-9}
    Method     &  Geo3k & MathVista & We-Math & MMK12 & MathVerse & LogicVista & MMMU$^\text{Pro}$ & Avg. \\ \cline{1-9}
    Teacher \tiny{\textit{Qwen3-VL-8B-Thinking}}  & 54.58&	65.20&	66.15&	42.55&	63.81&	43.40&	39.83&	53.65 \\
    Student \tiny{\textit{Qwen3-VL-2B-Thinking}} &  26.29&	43.10&	25.17&	13.00&	28.21&	18.57&	14.51&	24.12 \\
    \cline{1-9}
    Na\"ive Response Distillation &  35.94&	54.50&	51.38&	26.10&	48.67&	28.64&	22.60&	38.26 \\ \cdashline{1-9}[0.2pt/1pt]
    LLaVA-KD$^{\dagger}$~\cite{llava-kd} &  38.27 &	55.30&	56.32&	26.45&	51.10&	30.87&	24.05&	40.34 \\
    CompoDistill$^{\dagger}$~\cite{compodistill} &38.94&	57.50&	57.07&	28.30&	49.50&	34.80&	24.51&	41.52\\
    Align-TI$^{\dagger}$~\cite{align_kd} &  38.27&	56.60&	53.97&	27.95&	49.36&	33.33&	24.05&	40.50 \\ \cdashline{1-9}[0.2pt/1pt]
    {Masking-KD} (ours)  &  \textbf{40.93}&	\textbf{59.20}&	\textbf{63.79}&	\textbf{37.20}&	\textbf{57.89}&	\textbf{41.61}&	\textbf{30.75}&	\textbf{47.34} \\
    \hline
    \hline
    Undistilled \tiny{\textit{Qwen3-VL-4B-Thinking}} & 43.93&	62.60&	49.37&	31.55&	49.86&	39.37&	32.08 & 44.11\\
    \cline{1-9}
    \end{tabular}
    }
    \vspace{-0.35cm}
    \label{tab:main_vlm_distill_2b}
\end{table}
\begin{table}[t!]
    \centering
    \scriptsize
    \renewcommand{\arraystretch}{1.2}
    \setlength{\tabcolsep}{4.1pt}   
        \caption{\textbf{Extension to supervised fine-tuning (SFT).}
    The 2B student is trained on the 8B teacher-generated responses with one-hot labels (\ie, without the teacher's logits).
    }
    \vspace{0.1cm}
    \scalebox{1.055}{
    \begin{tabular}{m{1.7cm}|*{7}{>{\centering\arraybackslash}m{1.1cm}}|>{\centering\arraybackslash}m{1.1cm}}
    \cline{1-9}
    Method     &  Geo3k & MathVista & We-Math & MMK12 & MathVerse & LogicVista & MMMU$^\text{Pro}$ & Avg. \\ \cline{1-9}
    SFT & 35.77 & 52.60 & 48.39 & 23.85 & 45.00 & 26.85 & 19.88 & 36.05 \\ \cdashline{1-9}[0.2pt/1pt]
    SFT + ours & \textbf{40.60} & \textbf{55.00} & \textbf{57.99} & \textbf{33.40} & \textbf{54.50} & \textbf{34.45} & \textbf{26.01} & \textbf{43.14} \\
    \hline
    \end{tabular}
    }
    \vspace{-0.3cm}
    \label{tab:sft}
\end{table}

\begin{table}[t]
    \centering
    \scriptsize
    \renewcommand{\arraystretch}{1.2}
    \setlength{\tabcolsep}{4pt}   
    \newcommand{\rv}[2]{\Gape[2pt][1.5pt]{\makecell{#1\\[-1pt]{\tiny(#2)}}}}
        \caption{\textbf{Robustness to perturbed reasoning prefixes.}
    The student's reasoning prefixes are perturbed at every decoding step at inference by (a) replacing numerical values with incorrect ones or (b) randomly removing prefix tokens.
    }
    \label{tab:robustness}
    \vspace{0.1cm}
    \begin{subtable}[t]{0.49\linewidth}
        \centering
        \vspace{-0.2cm}
        \caption{Corrupted prefixes}
        \vspace{-0.1cm}
        \begin{tabular}{l!{\vrule}cccc}
        \cline{1-5}
        Corruption Ratio & Clean & 30\% & 50\% & 100\% \\ \cline{1-5}
        Na\"ive Distillation & 40.50 & \rv{37.25}{$-$8.0\%} & \rv{36.50}{$-$9.9\%} & \rv{33.50}{$-$17.3\%} \\ \cdashline{1-5}[0.2pt/1pt]
        Masking-KD (ours) & \textbf{53.00} & \rv{\textbf{52.50}}{$-$0.9\%} & \rv{\textbf{50.00}}{$-$5.7\%} & \rv{\textbf{47.75}}{$-$9.9\%} \\
        \hline
        \end{tabular}
        \label{tab:robustness_a}
    \end{subtable}
    \hfill
    \begin{subtable}[t]{0.49\linewidth}
        \centering
        \vspace{-0.2cm}
        \caption{Dropped prefixes}
        \vspace{-0.1cm}
        \begin{tabular}{l!{\vrule}cccc}
        \cline{1-5}
        Drop Ratio & Clean & 10\% & 30\% & 50\% \\ \cline{1-5}
        Na\"ive Distillation & 40.50 & \rv{39.00}{$-$3.7\%} & \rv{38.50}{$-$4.9\%} & \rv{34.25}{$-$15.4\%} \\ \cdashline{1-5}[0.2pt/1pt]
        Masking-KD (ours) & \textbf{53.00} & \rv{\textbf{53.00}}{0.0\%} & \rv{\textbf{51.75}}{$-$2.4\%} & \rv{\textbf{48.50}}{$-$8.5\%} \\
        \hline
        \end{tabular}
        \label{tab:robustness_b}
    \end{subtable}
    \vspace{-0.35cm}
\end{table}
\begin{table}[t]
    \centering
    \scriptsize
    \renewcommand{\arraystretch}{1.2}
    \setlength{\tabcolsep}{4pt}   
    \caption{\textbf{Ablation on each proposed method.} We begin with the na\"ive response distillation.
    Extensive ablation studies are provided in Appendix~\ref{appendix:additional_ablation}.
    }
    \vspace{0.1cm}
    \scalebox{1.04}{
    \begin{tabular}{@{}c|c|*{7}{>{\centering\arraybackslash}m{0.95cm}}|>{\centering\arraybackslash}m{0.65cm}@{\hspace{0.5pt}}}
    \hline
    \multirow{3}{*}{\makecell{Token-wise\\Salient Reasoning-\\prefix Masking}}  & \multirow{3}{*}{\makecell{Self-Paced \\Masking Budget\\Scheduling}}   &  \multirow{3}{*}{Geo3k} & \multirow{3}{*}{MathVista} & \multirow{3}{*}{We-Math} & \multirow{3}{*}{MMK12} & \multirow{3}{*}{MathVerse} & \multirow{3}{*}{LogicVista} & \multirow{3}{*}{MMMU$^\text{Pro~}$} & \multirow{3}{*}{Avg.} \\ 
    & & & & & & & & & \\
     & & & & & & & & & \\
    \hline
     & & 35.94&	54.50&	51.38&	26.10&	48.67&	28.64&	22.60&	38.26 \\
      \checkmark & & 40.03&	57.50&	62.46&	35.70&	57.75&	38.70&	29.23&	45.91 \\
     \checkmark  & \checkmark & \textbf{40.93}&	\textbf{59.20}&	\textbf{63.79}&	\textbf{37.20}&	\textbf{57.89}&	\textbf{41.61}&	\textbf{30.75}&	\textbf{47.34}\\
    \hline
    \end{tabular}
    }
    \vspace{-0.45cm}
    \label{tab:each_method}
\end{table}

\vspace{-0.25cm}
\subsection{Ablation Study}
\vspace{-0.25cm}
We conduct ablation studies using the 8B teacher and 2B student from Qwen3-VL-Thinking models.
\vspace{-0.28cm}

\paragraph{Effects of Each Proposed Method.}
In Tab.~\ref{tab:each_method}, we analyze the contribution of each method in our framework.
Starting with the na\"ive response distillation, where the student is distilled from teacher-generated responses using reverse KL divergence and the causal mask.
Guiding the student with our salient reasoning-prefix mask (Sec.~\ref{subsec:token-wise}) with a static cumulative ratio threshold, yields significant performance improvements.
This result demonstrates that masking salient prefixes leads to more effective reasoning distillation.
Introducing self-paced masking budget scheduling (Sec.~\ref{subsec:self-paced}) further enhances performance by adaptively controlling the masking scale for each distilled token based on its distillation difficulty.
This suggests that easily imitated tokens benefit from stronger masking, as they otherwise provide weak learning signals, while more difficult tokens require greater access to thought context for stable distillation.
Extensive ablation studies are provided in Appendix~\ref{appendix:additional_ablation}.

\vspace{-0.15cm}
\paragraph{Ablation on Masked Region.}
\begin{wraptable}{r}{0.25\textwidth}
    \vspace{-0.3cm}
    \centering
    \footnotesize
    \renewcommand{\arraystretch}{1.2}
    \setlength{\tabcolsep}{4pt}   
    \caption{Masked regions.}
    \vspace{-0.2cm}
    \begin{tabular}{l|c}
    \hline
    Masked Region     &  Avg. \\
    \hline
    Visual tokens      &  37.39 \\
    Question tokens & 31.31 \\ \cdashline{1-2}[0.2pt/1pt]
    Response (ours) & \textbf{47.34} \\
    \hline
    \hline
    Na\"ive Distillation & 38.26 \\
    \hline
    \end{tabular}
    \label{tab:ablation_masked_regions}
    \vspace{-0.2cm}
\end{wraptable}
Our methods apply the masking to prefixed response tokens. 
In this ablation (Tab.~\ref{tab:ablation_masked_regions}), we study different masking regions, such as visual tokens or question tokens, to validate the effectiveness of our response-prefix masking.
For visual and question masking, we extract response-to-visual or response-to-question attention from the auxiliary student forward, instead of response-to-response attention, and apply the same top-$\rho$ masking (Eq.~\eqref{eq:cumulative}) within the selected region.
All other components (\eg, self-paced masking budget scheduling and hyperparameters) remain unchanged.
The response-prefix masking achieves the best results, whereas the masking of visual or question tokens performs worse than na\"ive response distillation.
In contrast, our reasoning-prefix masking suppresses exposed textual cues, encouraging the student to draw on other available information.
Further analysis on these matters is discussed in Sec.~\ref{subsec:effect_masking}.


\begin{figure}[t]
    \centering

\includegraphics[width=0.97\linewidth]{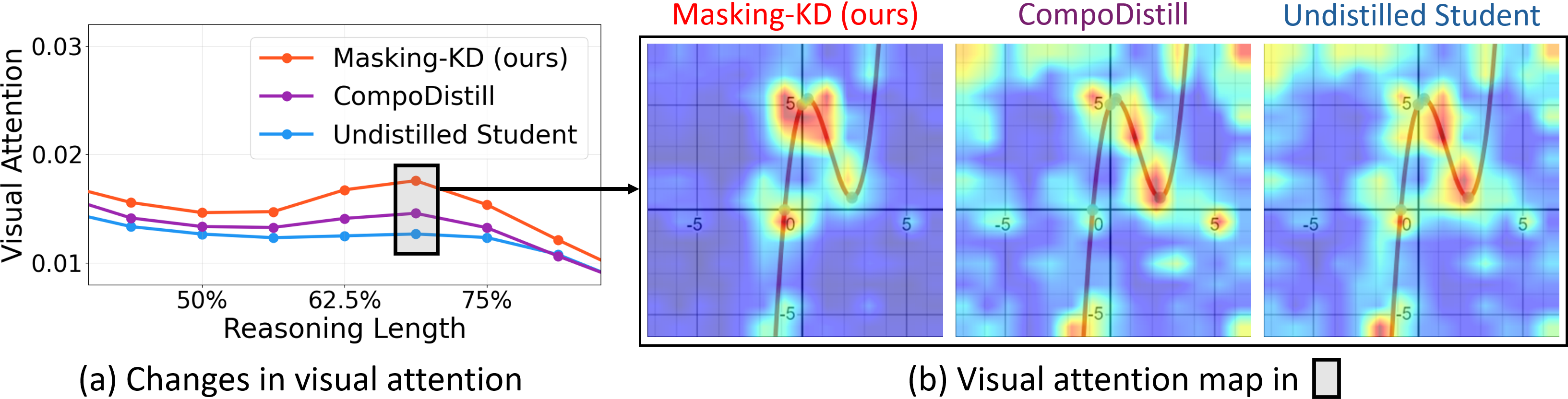}
\vspace{-0.1cm}
    \caption{\textbf{Evidence on visual-anchored thinking.}
    (a) changes in visual attention as generation proceeds, and (b) an example of visual attention maps at the peak attention point in \colorbox{gray!30}{gray box}. 
    }
    \label{fig:visual_forgetting_warp}
    \vspace{-0.1cm}
\end{figure}

\begin{figure}[t]
    \centering

\includegraphics[width=\linewidth]{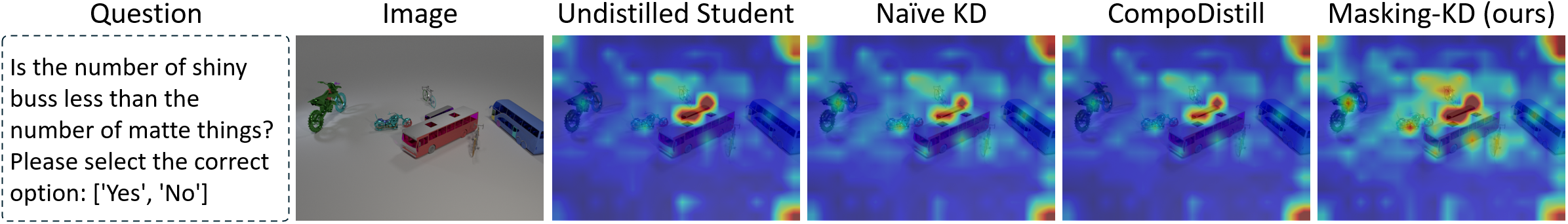}
    \caption{\textbf{Comparison on visual attention map.}
    We average the visual attention scores over the entire thinking trace. More visualizations are present in Fig.~\ref{fig:app_overall_visual_attn} of Appendix~\ref{app:subsec:visual_attn_comp}.
    }
    \label{fig:qual_overall}
    \vspace{-0.4cm}
\end{figure}

\vspace{-0.3cm}
\subsection{Effect of Salient Reasoning-prefix Masking}
\label{subsec:effect_masking}
\vspace{-0.2cm}

\paragraph{Visual-anchored Thinking.}
To evaluate whether our method improves the student’s use of visual information during reasoning, we compare the visual attention ratio over generation with the undistilled student (\ie, Qwen3-VL-2B-Thinking~\cite{qwen3_vl}) and CompoDistill~\cite{compodistill} in Fig.~\ref{fig:visual_forgetting_warp}a.
We also visualize the attention maps at the peak attention point in Fig.~\ref{fig:visual_forgetting_warp}b.
For robust analysis, we average the results over 10K responses generated by each model.
The undistilled student shows a rapid decline in visual attention, indicating visual forgetting~\cite{more_thought_less_acc}, while CompoDistill only partially mitigates this degradation.
In contrast, our method maintains the highest visual attention throughout generation, showing its effectiveness in alleviating visual forgetting and promoting visual-anchored thinking.


\vspace{-0.25cm}
\paragraph{Comparison on Visual Attention Map.}
In Fig.~\ref{fig:qual_overall}, we compare the visual attention maps over the entire thinking trace with the undistilled student, na\"ive KD, and CompoDistill~\cite{compodistill} to highlight our superiority in enhancing the visual perception ability of the student.
To obtain these visual attention maps, we average the visual attention scores over the entire thinking trace.
Although CompoDistill directly distills the teacher's visual attention maps, Masking-KD attends more strongly to relevant image regions.
This shows that our method more effectively encourages the student to exploit visual evidence during reasoning.
More visualizations are provided in Appendix~\ref{app:subsec:visual_attn_comp}.

\begin{figure}[t]
    \centering

\includegraphics[width=\linewidth]{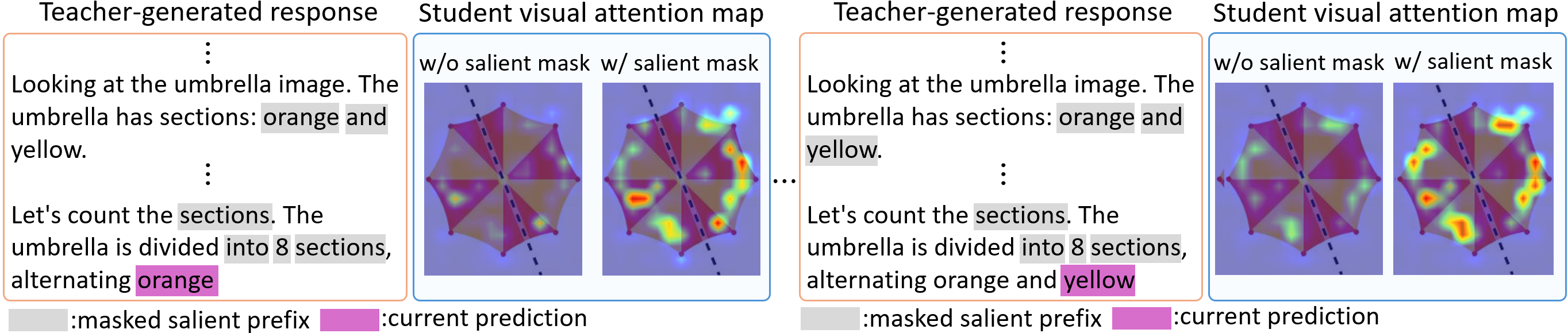}
    \caption{\textbf{Prediction behavior of the student during distillation} without and with our salient reasoning-prefix mask.
    Without a salient mask, the student uses a standard causal mask to predict the \colorbox{purple!40}{current token}\vspace{-0.4mm}.
    With a salient mask, the student exploits more visual information to compensate for the \colorbox{gray!20}{masked salient reasoning prefix}\vspace{-0.4mm}.
    More visualizations are provided in Fig.~\ref{fig:app_qual} of Appendix~\ref{app:subsec:predict_behavior}.
    }
    \label{fig:qual_distill}
    \vspace{-0.1cm}
\end{figure}

\vspace{-0.25cm}
\paragraph{Exploiting Visual Evidence during Distillation.}
In Fig.~\ref{fig:qual_distill}, we qualitatively compare the prediction behavior of the student with and without our salient reasoning-prefix masking.
The figure visualizes which response prefixes are masked at a given decoding step and how the visual attention map is activated when predicting the current token. 
Without salient masking, the student relies on salient textual prefixes in the response, resulting in relatively weak attention to the image. 
In contrast, our masking strategy removes these highly influential prefixes, leading the student to compensate for them by exploiting visual evidence. 
This leads to stronger activation in relevant image regions during distillation, encouraging that salient reasoning-prefix masking improves the student's visual perception throughout its thinking process.
More visualizations are provided in Appendix~\ref{app:subsec:predict_behavior}.

\begin{figure}[t]
    \centering
    \begin{subfigure}{0.45\linewidth}
        \centering
        \includegraphics[width=\linewidth]{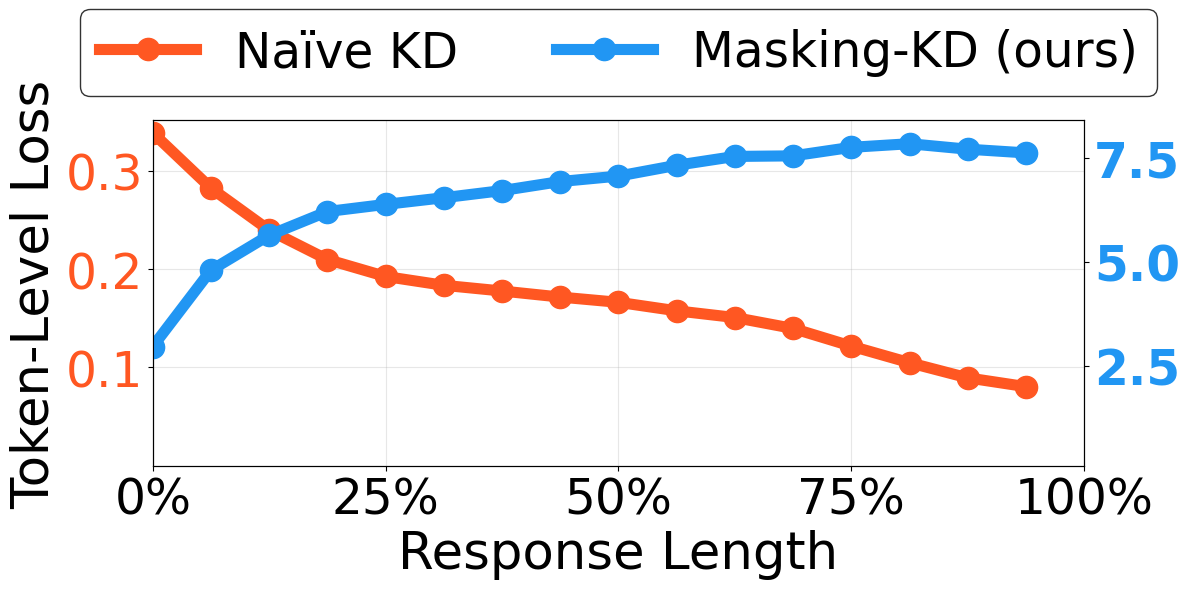}
        \vspace{-0.45cm}
        \caption{Decay of Reverse KL Divergence.}
        \label{fig:reverse_kl_div}
    \end{subfigure}%
    \hspace{0.04\linewidth}
    \begin{subfigure}{0.45\linewidth}
        \centering
        \includegraphics[width=\linewidth]{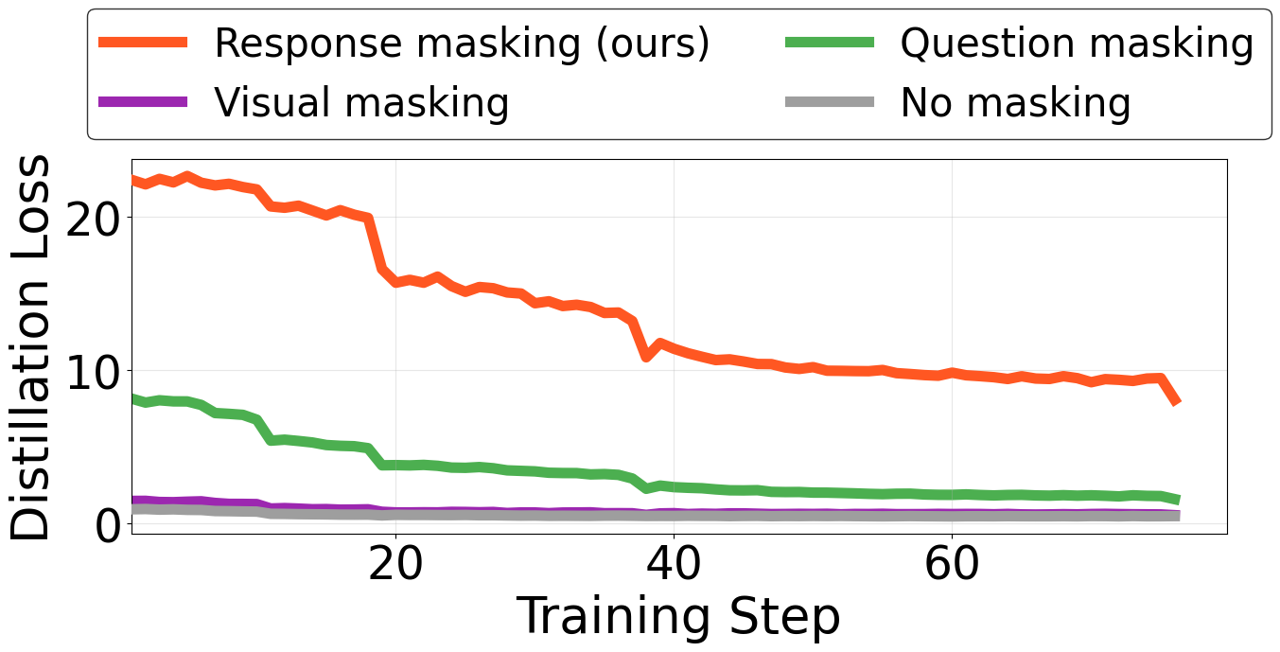}
        \vspace{-0.45cm}
        \caption{Distillation Loss Comparison.}
        \label{fig:distillation_loss}
    \end{subfigure}
    \vspace{-0.1cm}
    \caption{\textbf{Evidence on textual shortcut learning.} (a) Reverse KL divergence along the response. (b) Training loss when masking different regions as in Tab.~\ref{tab:ablation_masked_regions}, whose scale differs from (a) due to the distillation temperature $\tau$.}
    \label{fig:textual_shortcut}
    \vspace{-0.5cm}
\end{figure}

\vspace{-0.25cm}
\paragraph{Evidence on Textual Shortcut Learning.}
In Fig.~\ref{fig:reverse_kl_div}, we analyze the reverse KL divergence along the response, averaged over 19K teacher-generated responses, to examine whether the student relies on exposed reasoning prefixes during distillation.
Under na\"ive distillation, the reverse KL divergence gradually decreases as reasoning prefixes accumulate, indicating that the student can easily imitate subsequent tokens from exposed textual cues, whereas it increases under Masking-KD as these cues are masked.
We further compare the training loss when masking different regions in Fig.~\ref{fig:distillation_loss}, where each region is masked with the same procedure as in Tab.~\ref{tab:ablation_masked_regions}.
Masking visual tokens barely changes the loss from na\"ive distillation, while masking response prefixes yields the highest loss, showing that the student relies heavily on response prefixes when imitating the teacher.
Note that the reverse KL divergence in Fig.~\ref{fig:reverse_kl_div} is measured without the distillation temperature, whereas the training loss in Fig.~\ref{fig:distillation_loss} is computed with the distillation temperature $\tau$ during distillation, resulting in their different scales.

\begin{tcolorbox}[colback=gray!5,colframe=black,title=Takeaway: our key findings]
\begin{itemize}[leftmargin=2pt]
    \vspace{-0.1cm}
    \item When distilling long traces from think-answer VLMs, gradually accumulated reasoning prefixes offer sufficient information, allowing the student to easily imitate the teacher through exposed textual cues, reducing the need to refer to visual evidence (Fig.~\ref{fig:qual_distill} and Fig.~\ref{fig:textual_shortcut}).
    \item Masking salient reasoning prefixes during distillation encourages the student to exploit more visual information to compensate for missing textual cues (Fig.~\ref{fig:qual_distill}), improving the student's visual-anchored thinking (Fig.~\ref{fig:visual_forgetting_warp} and Fig.~\ref{fig:qual_overall}). 
    \item Rather than explicitly transferring the teacher's visual knowledge, guiding the student itself to anchor its own reasoning in visual information is effective for VLM reasoning distillation. (Tab.~\ref{tab:main_vlm_distill_4b} and Tab.~\ref{tab:main_vlm_distill_2b}). 
    \vspace{-0.1cm}
\end{itemize}
\end{tcolorbox}

\vspace{-0.25cm}
\section{Related Work}
\label{sec:related}

\vspace{-0.3cm}
\paragraph{Think-answer Reasoning.}
Recent think-answer models, such as DeepSeek-R1~\cite{deepseek-r1} and OpenAI o1~\cite{openai_o1}, have demonstrated that explicitly generating intermediate reasoning before the final answer can substantially improve performance on complex reasoning tasks~\cite{math500, mmlu}.
This think-answer paradigm has recently been extended to vision-language models (VLMs) to enable longer and more deliberate reasoning trajectories for multimodal problem solving.
For example, VL-Rethinker~\cite{rethinker} strengthens slow-thinking behavior through reinforcement learning~\cite{grpo} and explicit rethinking.
R1-OneVision~\cite{R1_onevision} bridges visual perception and deep reasoning through cross-modal formalization and step-wise reasoning supervision.
However, larger think-answer VLMs introduce high computational costs, motivating knowledge distillation into compact VLMs, which we explore in this work.

\vspace{-0.25cm}
\paragraph{Knowledge Distillation in VLMs.}
Knowledge distillation (KD) is widely used to compress vision-language models (VLMs) by transferring capabilities from a larger teacher to a smaller student.
Existing VLM distillation methods~\cite{align_kd, switch_kd, move_kd, llava-kd, compodistill, align_kd_visual} typically focus on preserving the teacher's visual knowledge through intermediate supervision, such as visual relation distillation in LLaVA-KD~\cite{llava-kd}, visual attention alignment in CompoDistill~\cite{compodistill}, and instruction-aware visual focus in Align-TI~\cite{align_kd}.
While effective for conventional VLM distillation, these methods do not address a challenge unique to think-answer VLMs: as long reasoning traces unfold, accumulated reasoning prefixes become highly informative, allowing the student to imitate the teacher through exposed textual cues rather than sustaining visually anchored reasoning.
{To address this, we propose a first think-answer distillation framework that suppresses shortcut textual cues, encouraging the student to rely on the remaining multimodal evidence by masking prefixed salient reasoning cues.}

\vspace{-0.25cm}
\paragraph{Self-Distillation.}
Self-distillation has emerged as an effective paradigm for improving model performance without a separate, larger teacher.
In LLMs, OPSD~\cite{opsd} uses the student's own reasoning trace as input to a self-teacher, which then provides feedback to improve the student.
In VLMs, SDRT~\cite{sdrt} self-distills from diverse reasoning traces generated by the model itself, training the student to imitate multiple reasoning paths.
While effective, these methods often require extra prompt design and multi-trace construction to create self-feedback signals.
In contrast, our approach simply masks salient reasoning prefixes for the student, while the self-teacher observes the full reasoning context, encouraging the student to exploit visual evidence to recover the missing textual cues.

\vspace{-0.4cm}
\section{Conclusion}
\label{sec:conclusion}

\vspace{-0.3cm}

In this paper, we present Masking-KD, a novel think-answer VLM distillation framework that masks salient reasoning prefixes to prevent the student from over-relying on exposed textual cues during distillation.
By encouraging the student to exploit visual evidence for recovering missing reasoning context, our method promotes more visual-anchored thinking during generation. 
Extensive experiments and analyses demonstrate that Masking-KD achieves outstanding results on multimodal reasoning benchmarks and effectively alleviates visual forgetting, arised in think-answer VLMs.

\subsection*{Acknowledgements}
This work was supported by the AI Graduate School Program at POSTECH (RS-2019-II191906 (5\%)), the IITP grants (RS-2026-25518317 (50\%), RS-2026-25616370 (30\%)), and the InnoCORE program (N10250156 (15\%)) funded by MSIT,Korea.

{
\small
\bibliographystyle{plain}
\bibliography{reference}

@article{grpo,
  title={Deepseekmath: Pushing the limits of mathematical reasoning in open language models},
  author={Shao, Zhihong and Wang, Peiyi and Zhu, Qihao and Xu, Runxin and Song, Junxiao and Bi, Xiao and Zhang, Haowei and Zhang, Mingchuan and Li, YK and others},
  journal={arXiv preprint arXiv:2402.03300},
  year={2024}
}

@article{qwen3_vl,
  title={Qwen3-vl technical report},
  author={Bai, Shuai and Cai, Yuxuan and Chen, Ruizhe and Chen, Keqin and Chen, Xionghui and Cheng, Zesen and Deng, Lianghao and Ding, Wei and Gao, Chang and Ge, Chunjiang and others},
  journal={arXiv preprint arXiv:2511.21631},
  year={2025}
}

@misc{openai_gpt54,
  author = {{OpenAI}},
  title = {Introducing GPT-5.4},
  year = {2026},
  howpublished = {https://openai.com/index/introducing-gpt-5-4/}
}

@article{openai_o1,
  title={Openai o1 system card},
  author={Jaech, Aaron and Kalai, Adam and Lerer, Adam and Richardson, Adam and El-Kishky, Ahmed and Low, Aiden and Helyar, Alec and Madry, Aleksander and Beutel, Alex and Carney, Alex and others},
  journal={arXiv preprint arXiv:2412.16720},
  year={2024}
}

@misc{gemini_41,
  author = {{The Gemini Team}},
  title = {Gemini 3.1 Pro: A smarter model for your most complex tasks},
  year = {2026},
}

@inproceedings{papo,
  title={Perception-aware policy optimization for multimodal reasoning},
  author={Wang, Zhenhailong and Guo, Xuehang and Stoica, Sofia and Xu, Haiyang and Wang, Hongru and Ha, Hyeonjeong and Chen, Xiusi and Chen, Yangyi and Yan, Ming and Huang, Fei and others},
  booktitle={ICLR},
  year={2026}
}

@inproceedings{rethinker,
  title={VL-Rethinker: Incentivizing Self-Reflection of Vision-Language Models with Reinforcement Learning},
  author={Wang, Haozhe and Qu, Chao and Huang, Zuming and Chu, Wei and Lin, Fangzhen and Chen, Wenhu},
  booktitle={NeurIPS},
  year={2025}
}

@inproceedings{R1_onevision,
  title={R1-onevision: Advancing generalized multimodal reasoning through cross-modal formalization},
  author={Yang, Yi and He, Xiaoxuan and Pan, Hongkun and Jiang, Xiyan and Deng, Yan and Yang, Xingtao and Lu, Haoyu and Yin, Dacheng and Rao, Fengyun and Zhu, Minfeng and others},
  booktitle={ICCV},
  year={2025}
}

@article{recursive,
  title={Recursive Think-Answer Process for LLMs and VLMs},
  author={Lee, Byung-Kwan and Chee, Youngchae and Ro, Yong Man},
  journal={CVPR Findings},
  year={2026}
}

@article{r1_zero,
  title={R1-Zero's" Aha Moment" in Visual Reasoning on a 2B Non-SFT Model},
  author={Zhou, Hengguang and Li, Xirui and Wang, Ruochen and Cheng, Minhao and Zhou, Tianyi and Hsieh, Cho-Jui},
  journal={arXiv preprint arXiv:2503.05132},
  year={2025}
}

@article{deepseek-r1,
  title={Deepseek-r1: Incentivizing reasoning capability in llms via reinforcement learning},
  author={Guo, Daya and Yang, Dejian and Zhang, Haowei and Song, Junxiao and Wang, Peiyi and Zhu, Qihao and Xu, Runxin and Zhang, Ruoyu and Ma, Shirong and Bi, Xiao and others},
  journal={arXiv preprint arXiv:2501.12948},
  year={2025}
}

@inproceedings{pixel_reasoner,
  title={Pixel reasoner: Incentivizing pixel-space reasoning with curiosity-driven reinforcement learning},
  author={Wang, Haozhe and Su, Alex and Ren, Weiming and Lin, Fangzhen and Chen, Wenhu},
  booktitle={NeurIPS},
  year={2025}
}

@article{more_thought_less_acc,
  title={More thought, less accuracy? on the dual nature of reasoning in vision-language models},
  author={Tian, Xinyu and Zou, Shu and Yang, Zhaoyuan and He, Mengqi and Waschkowski, Fabian and Wesemann, Lukas and Tu, Peter and Zhang, Jing},
  journal={ICLR},
  year={2026}
}

@inproceddings{mmeureka,
  title={Mm-eureka: Exploring the frontiers of multimodal reasoning with rule-based reinforcement learning},
  author={Meng, Fanqing and Du, Lingxiao and Liu, Zongkai and Zhou, Zhixiang and Lu, Quanfeng and Fu, Daocheng and Han, Tiancheng and Shi, Botian and Wang, Wenhai and He, Junjun and others},
  journal={arXiv preprint arXiv:2503.07365},
  year={2025}
}

@article{compodistill,
  title={Compodistill: Attention distillation for compositional reasoning in multimodal llms},
  author={Kim, Jiwan and Kim, Kibum and Seo, Sangwoo and Park, Chanyoung},
  journal={ICLR},
  year={2026}
}

@article{align_kd_visual,
  title={Align-kd: Distilling cross-modal alignment knowledge for mobile vision-language model},
  author={Feng, Qianhan and Li, Wenshuo and Lin, Tong and Chen, Xinghao},
  journal={CVPR},
  year={2025}
}

@article{align_kd,
  title={Beyond Next-Token Alignment: Distilling Multimodal Large Language Models via Token Interactions},
  author={Chen, Lin and Zhao, Xiaoke and Ding, Kun and Feng, Weiwei and Miao, Changtao and Wang, Zili and Guo, Wenxuan and Wang, Ying and Zheng, Kaiyuan and Zhang, Bo and others},
  journal={arXiv preprint arXiv:2602.09483},
  year={2026}
}

@inproceedings{llava-kd,
  title={Llava-kd: A framework of distilling multimodal large language models},
  author={Cai, Yuxuan and Zhang, Jiangning and He, Haoyang and He, Xinwei and Tong, Ao and Gan, Zhenye and Wang, Chengjie and Xue, Zhucun and Liu, Yong and Bai, Xiang},
  booktitle={ICCV},
  year={2025}
}

@article{switch_kd,
  title={Switch-KD: Visual-Switch Knowledge Distillation for Vision-Language Models},
  author={Sun, Haoyi and Wang, Xiaoxiao and Mao, Ning and Wang, Qian and Mu, Lifu and Zheng, Wen and Wei, Tao and Chen, Wei},
  journal={CVPR Findings},
  year={2026}
}

@inproceedings{move_kd,
  title={Move-kd: Knowledge distillation for vlms with mixture of visual encoders},
  author={Cao, Jiajun and Zhang, Yuan and Huang, Tao and Lu, Ming and Zhang, Qizhe and An, Ruichuan and Ma, Ningning and Zhang, Shanghang},
  booktitle={CVPR},
  year={2025}
}

@article{sdrt,
  title={SDRT: Enhance Vision-Language Models by Self-Distillation with Diverse Reasoning Traces},
  author={Wu, Guande and Song, Huan and Wang, Yawei and Yan, Qiaojing and Tian, Yijun and Cheong, Lin Lee and Xu, Panpan},
  journal={arXiv preprint arXiv:2503.01754},
  year={2025}
}

@inproceedings{gkd,
  title={On-policy distillation of language models: Learning from self-generated mistakes},
  author={Agarwal, Rishabh and Vieillard, Nino and Zhou, Yongchao and Stanczyk, Piotr and Garea, Sabela Ramos and Geist, Matthieu and Bachem, Olivier},
  booktitle={ICLR},
  year={2024}
}

@article{opsd,
  title={Self-Distilled Reasoner: On-Policy Self-Distillation for Large Language Models},
  author={Zhao, Siyan and Xie, Zhihui and Liu, Mengchen and Huang, Jing and Pang, Guan and Chen, Feiyu and Grover, Aditya},
  journal={arXiv preprint arXiv:2601.18734},
  year={2026}
}

@article{attention_is_all,
  title={Attention is all you need},
  author={Vaswani, Ashish and Shazeer, Noam and Parmar, Niki and Uszkoreit, Jakob and Jones, Llion and Gomez, Aidan N and Kaiser, {\L}ukasz and Polosukhin, Illia},
  journal={NeurIPS},
  year={2017}
}

@inproceedings{mmlu,
  title={Measuring Massive Multitask Language Understanding},
  author={Hendrycks, Dan and Burns, Collin and Basart, Steven and Zou, Andy and Mazeika, Mantas and Song, Dawn and Steinhardt, Jacob},
  booktitle={ICLR},
  year={2021}
}

@inproceedings{math500,
  title={Measuring Mathematical Problem Solving With the MATH Dataset},
  author={Hendrycks, Dan and Burns, Collin and Kadavath, Saurav and Arora, Akul and Basart, Steven and Tang, Eric and Song, Dawn and Steinhardt, Jacob},
  booktitle={NeurIPS Datasets and Benchmarks},
  year={2021}
}

@inproceedings{mmmu_pro,
  title={Mmmu-pro: A more robust multi-discipline multimodal understanding benchmark},
  author={Yue, Xiang and Zheng, Tianyu and Ni, Yuansheng and Wang, Yubo and Zhang, Kai and Tong, Shengbang and Sun, Yuxuan and Yu, Botao and Zhang, Ge and Sun, Huan and others},
  booktitle={ACL},
  year={2025}
}

@inproceedings{mathvista,
  title={MathVista: Evaluating Mathematical Reasoning of Foundation Models in Visual Contexts},
  author={Lu, Pan and Bansal, Hritik and Xia, Tony and Liu, Jiacheng and Li, Chunyuan and Hajishirzi, Hannaneh and Cheng, Hao and Chang, Kai-Wei and Galley, Michel and Gao, Jianfeng},
  booktitle={ICLR},
  year={2024}
}

@inproceedings{wemath,
  title={We-math: Does your large multimodal model achieve human-like mathematical reasoning?},
  author={Qiao, Runqi and Tan, Qiuna and Dong, Guanting and MinhuiWu, MinhuiWu and Sun, Chong and Song, Xiaoshuai and Wang, Jiapeng and Gongque, Zhuoma and Lei, Shanglin and Zhang, Yifan and others},
  booktitle={ACL},
  year={2025}
}

@inproceedings{geo3k,
  title={Inter-gps: Interpretable geometry problem solving with formal language and symbolic reasoning},
  author={Lu, Pan and Gong, Ran and Jiang, Shibiao and Qiu, Liang and Huang, Siyuan and Liang, Xiaodan and Zhu, Song-Chun},
  booktitle={ACL},
  year={2021}
}

@inproceedings{mathverse,
  title={Mathverse: Does your multi-modal llm truly see the diagrams in visual math problems?},
  author={Zhang, Renrui and Jiang, Dongzhi and Zhang, Yichi and Lin, Haokun and Guo, Ziyu and Qiu, Pengshuo and Zhou, Aojun and Lu, Pan and Chang, Kai-Wei and Qiao, Yu and others},
  booktitle={ECCV},
  year={2024},
}

@inproceedings{logicvista,
  title={{LogicVista}: Multimodal llm logical reasoning benchmark in visual contexts},
  author={Xiao, Yijia and Sun, Edward and Liu, Tianyu and Wang, Wei},
  booktitle={ICLR},
  year={2025}
}

@inproceedings{science,
  title={Learn to explain: Multimodal reasoning via thought chains for science question answering},
  author={Lu, Pan and Mishra, Swaroop and Xia, Tanglin and Qiu, Liang and Chang, Kai-Wei and Zhu, Song-Chun and Tafjord, Oyvind and Clark, Peter and Kalyan, Ashwin},
  journal={NeurIPS},
  year={2022}
}

@article{ovis,
  title={Ovis: Structural embedding alignment for multimodal large language model, 2024},
  author={Lu, Shiyin and Li, Yang and Chen, Qing-Guo and Xu, Zhao and Luo, Weihua and Zhang, Kaifu and Ye, Han-Jia},
  journal={arXiv preprint arXiv:2405.20797},
  year={2024}
}

@article{mimo,
      title={MiMo-VL Technical Report}, 
      author={LLM-Core-Team Xiaomi},
      year={2025},
      journal={arXiv preprint arXiv:2506.03569}, 
}

@article{internvl3_5,
  title={InternVL3.5: Advancing Open-Source Multimodal Models in Versatility, Reasoning, and Efficiency},
  author={Wang, Weiyun and Gao, Zhangwei and Gu, Lixin and Pu, Hengjun and Cui, Long and Wei, Xingguang and Liu, Zhaoyang and Jing, Linglin and Ye, Shenglong and Shao, Jie and others},
  journal={arXiv preprint arXiv:2508.18265},
  year={2025}
}

@article{internvl3,
  title={Internvl3: Exploring advanced training and test-time recipes for open-source multimodal models},
  author={Zhu, Jinguo and Wang, Weiyun and Chen, Zhe and Liu, Zhaoyang and Ye, Shenglong and Gu, Lixin and Tian, Hao and Duan, Yuchen and Su, Weijie and Shao, Jie and others},
  journal={arXiv preprint arXiv:2504.10479},
  year={2025}
}

@article{qwen2_5,
  title={Qwen2. 5-VL Technical Report},
  author={Bai, Shuai and Chen, Keqin and Liu, Xuejing and Wang, Jialin and Ge, Wenbin and Song, Sibo and Dang, Kai and Wang, Peng and Wang, Shijie and Tang, Jun and others},
  journal={arXiv preprint arXiv:2502.13923},
  year={2025}
}

@inproceedings{top_p,
  title={The Curious Case of Neural Text Degeneration},
  author={Holtzman, Ari and Buys, Jan and Du, Li and Forbes, Maxwell and Choi, Yejin},
  booktitle={ICLR},
  year={2020}
}

@misc{spacy,
  author={Honnibal, Matthew and Montani, Ines and Van Landeghem, Sofie and Boyd, Adriane},
  title={{spaCy}: Industrial-strength Natural Language Processing in {Python}},
  howpublished={Zenodo},
  year={2020}
}

@inproceedings{pope,
  title={Evaluating Object Hallucination in Large Vision-Language Models},
  author={Li, Yifan and Du, Yifan and Zhou, Kun and Wang, Jinpeng and Zhao, Wayne Xin and Wen, Ji-Rong},
  booktitle={EMNLP},
  year={2023}
}
}
\newpage

\appendix

\begin{center}
\makeatletter
\@toptitlebar
\makeatother
    {\Large \textbf{Hide to See: Reasoning-prefix Masking for \\Visual-anchored Thinking in VLM Distillation} \\ - \textit{Appendix} - \par}
\makeatletter
    \@bottomtitlebar
\makeatother
\end{center}

\begin{center}
    \large{\textbf{Overview of Appendix}}
\end{center}
\newcommand{\appsubref}[2]{\hyperref[#1]{\textcolor{black}{#2}}}
We provide the table of contents for the Appendix below:
\begin{enumerate}[label=\Alph*.]
    \item \hyperref[appendix:additional_results]{Additional Experiments}
            \begin{enumerate}[label=\Alph{enumi}.\arabic*.]
            \item \appsubref{app:subsec:other_vlms}{Results on other VLM Models}
            \item \appsubref{app:subsec:pnp_kd}{Compatibility with other VLM Distillations}
            \item \appsubref{app:subsec:student_generated}{Results from Student-generated Response}
            \item \appsubref{app:subsec:computational}{Computational Comparison with other VLM Distillations}
            \item \appsubref{app:statistical}{Statistical Significance}
            \item \appsubref{app:subsec:pope}{Evaluation on Hallucination Benchmark}
        \end{enumerate}

    \item \hyperref[appendix:additional_ablation]{Additional Ablation Studies}
        \begin{enumerate}[label=\Alph{enumi}.\arabic*.]
            \item \appsubref{app:subsec:ablation_within}{Ablation Study within Proposed Method}
            \item \appsubref{app:subsec:auxiliary}{Auxiliary Student Forward}
            \item \appsubref{app:subsec:loss_functions}{Loss Functions}
            \item \appsubref{app:subsec:previous_token}{Excluding Immediate Previous Token from Masking}
        \end{enumerate}

    \item \hyperref[appendix:additional_analysis]{Additional Analyses}
        \begin{enumerate}[label=\Alph{enumi}.\arabic*.]
            \item \appsubref{app:subsec:trace_length}{Effectiveness on Longer Reasoning Traces}
            \item \appsubref{app:subsec:rho_analysis}{Analysis on Self-Paced Masking Budget}
            \item \appsubref{app:subsec:masked_position}{Statistics of Masked Prefix Positions}
            \item \appsubref{app:subsec:inference_example}{Example of Inference}
            \item \appsubref{app:subsec:visual_attn_comp}{More Comparison on Visual Attention Map}
            \item \appsubref{app:subsec:predict_behavior}{More Prediction Behavior of the Student during Distillation}
        \end{enumerate}

    \item \hyperref[appendix:implementation_details]{Additional Details}
        \begin{enumerate}[label=\Alph{enumi}.\arabic*.]
            \item \appsubref{app:subsec:top-rho}{Details on Top-$\rho$ Masking}
            \item \appsubref{app:subsec:self_distill}{Details on Self-Distillation}
            \item \appsubref{app:instruction}{Instruction for Teacher-generated Response}
        \end{enumerate}

    \item \hyperref[appendix:further_discussion]{Further Discussion}
        \begin{enumerate}[label=\Alph{enumi}.\arabic*.]
            \item \appsubref{appendix:limitation}{Limitation}
            \item \appsubref{appendix:social_impact}{Social Impact}
        \end{enumerate}
        

\end{enumerate}

\clearpage

\section{Additional Experiments}
\label{appendix:additional_results}
\vspace{-0.2cm}

\subsection{Results on other VLM Models}
\label{app:subsec:other_vlms}
\vspace{-0.2cm}
In this section, we validate the generality of our approach by applying it to other VLM models across both knowledge distillation and self-distillation settings.
We report results for InternVL3.5~\cite{internvl3_5} in Tab.~\ref{tab:app_other_vlm_intern} and Qwen3-VL-Instruct~\cite{qwen3_vl} in Tab.~\ref{tab:app_other_vlm_instruct}.
Masking-KD achieves the best performance across various teacher-student configurations (\ie, 8B--8B self-distillation, 8B--4B knowledge distillation, and 8B--2B knowledge distillation), demonstrating its generality across different VLM models.


\begin{table}[h]
    \vspace{-0.2cm}
    \centering
    \scriptsize
    \renewcommand{\arraystretch}{1.2}
    \setlength{\tabcolsep}{4.1pt}   
        \caption{\textbf{Results on other VLM models (InternVL3.5~\cite{internvl3_5}).}
        $\dagger$ denotes a self-distilled model that uses its own predictions under our salient reasoning-prefix mask, and $\ddagger$ indicates the distilled student from the 8B teacher. 
        }
        \vspace{0.1cm}
    \scalebox{1.055}{
    \begin{tabular}{m{3.2cm}|*{7}{>{\centering\arraybackslash}m{0.94cm}}|>{\centering\arraybackslash}m{0.7cm}}
    \cline{1-9}
    Method     &  Geo3k & MathVista & We-Math & MMK12 & MathVerse & LogicVista & MMMU$^\text{Pro}$ & Avg. \\ \cline{1-9}
    Teacher \tiny{\textit{InternVL3.5-8B}}  & 44.59&	68.50&	56.61&	44.95	&53.26&	37.81&	38.50	&49.17 \\  \cdashline{1-9}[0.2pt/1pt]
    \textit{Self-distill} Masking-KD$^\dagger$ (ours) & \textbf{46.26}&	\textbf{69.70}&	\textbf{58.79}&	\textbf{45.45}&	\textbf{54.97}&	\textbf{39.16}&	\textbf{39.84}&	\textbf{50.60} \\
    \hline
    \hline
    Student \tiny{\textit{InternVL3.5-4B}} &  41.93&	52.10&	45.46&	25.80	&43.21	&27.29&	29.77&	37.94\\ \cdashline{1-9}[0.2pt/1pt]
    Na\"ive Response Distillation & \textbf{44.93}&	52.30	&47.64	&27.60&	43.58&	27.74	&25.09&	38.41\\
    Masking-KD$^\ddagger$ (ours) & 44.26	&\textbf{56.40}&	\textbf{50.75}&	\textbf{30.45}&	\textbf{47.61}&	\textbf{30.20}&	\textbf{27.23}&	\textbf{40.99}\\
    \hline
    \hline
    Student \tiny{\textit{InternVL3.5-2B}} &  29.95	&40.60	&24.31&	14.70&	27.94&	16.11&	12.54&	23.74\\   \cdashline{1-9}[0.2pt/1pt]
        Na\"ive Response Distillation & 32.95&	43.50&	27.59	&13.40	&30.87&	18.12&	13.53&	25.71 \\
    Masking-KD$^\ddagger$ (ours) & \textbf{35.61}	&\textbf{52.00}	&\textbf{41.95}&	\textbf{19.70}&	\textbf{38.67}&	\textbf{25.73}&	\textbf{20.46}&	\textbf{33.45}\\

    \cline{1-9}
    
    \end{tabular}
    }
    \vspace{-0.4cm}
    \label{tab:app_other_vlm_intern}
\end{table}
\begin{table}[h]
    \vspace{-0.2cm}
    \centering
    \scriptsize
    \renewcommand{\arraystretch}{1.2}
    \setlength{\tabcolsep}{4.1pt}   
        \caption{\textbf{Results on other VLM models (Qwen3-VL-Instruct~\cite{qwen3_vl}).}
        $\dagger$ denotes a self-distilled model that uses its own predictions under our salient reasoning-prefix mask, and $\ddagger$ indicates the distilled student from the 8B teacher. 
    }
    \vspace{0.1cm}
    \scalebox{1.055}{
    \begin{tabular}{m{3.2cm}|*{7}{>{\centering\arraybackslash}m{0.94cm}}|>{\centering\arraybackslash}m{0.7cm}}
    \cline{1-9}
    Method     &  Geo3k & MathVista & We-Math & MMK12 & MathVerse & LogicVista & MMMU$^\text{Pro}$ & Avg. \\ \cline{1-9}
    Teacher \tiny{\textit{Qwen3-VL-8B-Instruct}}  &54.58	&67.40&	70.11&	58.20&	63.49&	54.36&	38.90&	58.15  \\ \cdashline{1-9}[0.2pt/1pt]
    \textit{Self-distill} Masking-KD$^\dagger$ (ours) &\textbf{ 59.23}&	\textbf{67.70}&	\textbf{74.48}	& \textbf{65.00}	&\textbf{68.35}	&\textbf{55.26}&	\textbf{44.80}&	\textbf{62.12}\\
    \hline
    \hline
    Student \tiny{\textit{Qwen3-VL-4B-Instruct}} & 51.91&	65.10&	58.28&	46.10&	45.05&	45.64&	19.83&	47.42 \\ \cdashline{1-9}[0.2pt/1pt]
    Na\"ive Response Distillation & 48.59	&\textbf{64.90}&	64.14&	48.00&	57.61&	47.43&	36.42&	52.44\\
    Masking-KD$^\ddagger$ (ours) & \textbf{54.74}&	64.60&	\textbf{65.98}&	\textbf{57.70}&	\textbf{61.06}&	\textbf{51.45}&	\textbf{40.12}&	\textbf{56.52}\\
    \hline
    \hline
    Student \tiny{\textit{Qwen3-VL-2B-Instruct}} &  26.46	&55.70&	33.56&	25.60&	30.00&	27.07&	20.46	&31.26 \\  \cdashline{1-9}[0.2pt/1pt]
        Na\"ive Response Distillation & 31.78	&57.10&	52.18	&33.05&	42.66	&34.00&	27.05&	39.69\\ 
    Masking-KD$^\ddagger$ (ours) & \textbf{36.77}&	\textbf{58.00}&	\textbf{55.52}	&\textbf{44.15}&	\textbf{48.67}	&\textbf{43.18}&	\textbf{30.29}&	\textbf{45.23}\\
    \cline{1-9}
    
    \end{tabular}
    }
    \vspace{-0.3cm}
    \label{tab:app_other_vlm_instruct}
\end{table}

\subsection{Compatibility with other VLM Distillations}
\label{app:subsec:pnp_kd}
\vspace{-0.2cm}
In Tab.~\ref{tab:pnp_kd}, we report the performance of Masking-KD when combined with other VLM distillation methods, including LLaVA-KD~\cite{llava-kd}, CompoDistill~\cite{compodistill}, and Align-TI~\cite{align_kd}.
Our approach consistently improves performance across these methods.
However, the performance gains are smaller when combined with na\"ive response distillation.
We attribute this to a partial conflict between prior methods that explicitly distill the teacher's visual knowledge, and our objective for enhancing the student's own use of visual evidence.
Nevertheless, the consistent improvements indicate our effectiveness in VLM distillations.
\begin{table}[h!]
    \vspace{-0.2cm}
    \centering
    \scriptsize
    \renewcommand{\arraystretch}{1.2}
    \setlength{\tabcolsep}{5pt}
     \caption{\textbf{Compatibility with other VLM distillations.} We adapt our salient reasoning-prefix masking into other VLM distillation methods.}
     \vspace{0.1cm}
    \scalebox{1.055}{
    \begin{tabular}{l|ccccccc|c}
    \cline{1-9}
    Method     &  Geo3k & MathVista & We-Math & MMK12 & MathVerse & LogicVista & MMMU$^\text{Pro}$ & Avg. \\ \cline{1-9}
    
    Na\"ive Response Distillation & 
\textcolor{gray}{35.94} & 
\textcolor{gray}{54.50} & 
\textcolor{gray}{51.38} & 
\textcolor{gray}{26.10} & 
\textcolor{gray}{48.67} & 
\textcolor{gray}{28.64} & 
\textcolor{gray}{22.60} & 
\textcolor{gray}{38.26} \\
    ~~w/ Masking-KD (ours) &  40.93&	59.20&	63.79&	37.20&	57.89&	41.61&	30.75&	47.34 \\
    \hline
    
    LLaVA-KD~\cite{llava-kd}&
 \textcolor{gray}{38.27} &	\textcolor{gray}{55.30}&	\textcolor{gray}{56.32}&	\textcolor{gray}{26.45}&	\textcolor{gray}{51.10}&	\textcolor{gray}{30.87}&	\textcolor{gray}{24.05}&	\textcolor{gray}{40.34} \\   
    ~~w/ Masking-KD (ours)  &43.09&	58.40&	64.43	&36.20	&57.75&	37.14&	29.31	&46.62	 \\
    \cline{1-9}
    
    CompoDistill~\cite{compodistill}
&\textcolor{gray}{38.94}&	\textcolor{gray}{57.50}&	\textcolor{gray}{57.07}&	\textcolor{gray}{28.30}&	\textcolor{gray}{49.50}&	\textcolor{gray}{34.80}&	\textcolor{gray}{24.51}&	\textcolor{gray}{41.52} \\   
    ~~w/ Masking-KD (ours) & 42.60&	58.90&	60.63	&32.35	&55.96&	33.46&	27.86&	44.54 \\
    \cline{1-9}
    
    Align-TI~\cite{align_kd} & \textcolor{gray}{38.27}&	\textcolor{gray}{56.60}&	\textcolor{gray}{53.97}&	\textcolor{gray}{27.95}&	\textcolor{gray}{49.36}&	\textcolor{gray}{33.33}&	\textcolor{gray}{24.05}&	\textcolor{gray}{40.50} \\  
    ~~w/ Masking-KD (ours) & 42.43 &	58.20&	56.69&	30.15	&53.16	&33.89	&25.97&	42.93 \\
    \hline
    \end{tabular}
    }
    \vspace{-0.3cm}
    \label{tab:pnp_kd}
\end{table}

\subsection{Results from Student-generated Response}
\label{app:subsec:student_generated}
\vspace{-0.2cm}
In the main manuscript, we conduct distillation using teacher-generated responses.
Here, we further show that our approach is also effective when distilling from student-generated responses, compared with other VLM distillations~\cite{compodistill, align_kd, llava-kd} in Tab.~\ref{tab:app_student_gen_4b} (8B teacher -- 4B student) and Tab.~\ref{tab:app_student_gen_4b} (8B teacher -- 2B student).
Using student-generated responses has the advantage of providing supervision that is closer to the student's own reasoning distribution, as described in \cite{gkd}.
This can reduce the distribution mismatch between training traces and the student's generation behavior, leading to more stable and better-aligned distillation.
These results demonstrate that our approach remains effective even when distilling from student-generated responses.

\begin{table}[h!]

    \vspace{-0.3cm}
    \centering
    \scriptsize
    \renewcommand{\arraystretch}{1.2}
    \setlength{\tabcolsep}{4.1pt}   
        \caption{\textbf{Distillation Results from Student-generated Responses (8B teacher -- 4B student).}
    For distillation, we use student-generated responses instead of teacher-generated responses, where the mismatched distribution between training traces and the student's generation behavior is alleviated. 
    }
    \vspace{0.1cm}
    \scalebox{1.055}{
    \begin{tabular}{m{3.2cm}|*{7}{>{\centering\arraybackslash}m{0.94cm}}|>{\centering\arraybackslash}m{0.7cm}}
    \cline{1-9}
    Method     &  Geo3k & MathVista & We-Math & MMK12 & MathVerse & LogicVista & MMMU$^{\text{Pro}}$ & Avg. \\ \cline{1-9}
    Teacher \tiny{\textit{Qwen3-VL-8B-Thinking}}  & 54.58&	65.20&	66.15&	42.55&	63.81&	43.40&	39.83&	53.65 \\
    Student \tiny{\textit{Qwen3-VL-4B-Thinking}} & 43.93&	62.60&	49.37&	31.55&	49.86&	39.37&	32.08 & 44.11\\
    \cline{1-9}
    Na\"ive Response Distillation & 48.92&	61.70&	58.91&	37.25&	57.52&	42.95&	33.64&	48.70 \\ 
    \cdashline{1-9}[0.2pt/1pt]
    LLaVA-KD$^{\dagger}$~\cite{llava-kd} &  49.58&	63.50&	60.86&	36.95&	57.89	&42.51&	33.64	&49.28 \\
    CompoDistill$^{\dagger}$~\cite{compodistill} & 49.25&	62.00&	61.21&	38.15&	58.21	&44.07&	33.82&	49.53 \\
    Align-TI$^{\dagger}$~\cite{align_kd} & 51.08&	61.70&	61.84&	38.50&	57.34	&42.73&	34.68&	49.70 \\
    \cdashline{1-9}[0.2pt/1pt]
    Masking-KD (ours)  & \textbf{54.91}&	\textbf{66.10}&	\textbf{69.43}&	\textbf{51.60}&	\textbf{63.94}&	\textbf{53.02}&	\textbf{40.52}&	\textbf{57.07} \\
    
    \hline
    \end{tabular}
    }
    \label{tab:app_student_gen_4b}
    \vspace{-0.5cm}
\end{table}

\begin{table}[h!]
    \vspace{-0.2cm}
    \centering
    \scriptsize
    \renewcommand{\arraystretch}{1.2}
    \setlength{\tabcolsep}{4.1pt}   
    \caption{\textbf{Distillation Results from Student-generated Responses (8B teacher -- 2B student).}
    }
    \vspace{0.1cm}
    \scalebox{1.055}{
    \begin{tabular}{m{3.2cm}|*{7}{>{\centering\arraybackslash}m{0.94cm}}|>{\centering\arraybackslash}m{0.7cm}}
    \cline{1-9}
    Method     &  Geo3k & MathVista & We-Math & MMK12 & MathVerse & LogicVista & MMMU$^\text{Pro}$ & Avg. \\ \cline{1-9}
    Teacher \tiny{\textit{Qwen3-VL-8B-Thinking}}  & 54.58&	65.20&	66.15&	42.55&	63.81&	43.40&	39.83&	53.65 \\
    Student \tiny{\textit{Qwen3-VL-2B-Thinking}} &  26.29&	43.10&	25.17&	13.00&	28.21&	18.57&	14.51&	24.12 \\
    \cline{1-9}
    Na\"ive Response Distillation &  32.45&	46.80&	41.55&	18.35&	40.14&	23.71&	16.36&	31.34 \\ \cdashline{1-9}[0.2pt/1pt]
    LLaVA-KD$^{\dagger}$~\cite{llava-kd} &  33.11 &	50.60	&44.05&	18.85&	42.32&	21.58&	17.75&	32.61 \\
    CompoDistill$^{\dagger}$~\cite{compodistill} &  33.44	&50.00	&45.75&	19.85&	43.21	&23.94	&18.61	&33.54 \\
    Align-TI$^{\dagger}$~\cite{align_kd} & 33.04	&49.51&	43.49	&18.65&	41.60	&22.82	&18.54&	32.52 \\
    \cdashline{1-9}[0.2pt/1pt]
    Masking-KD (ours) & \textbf{44.59}&	\textbf{58.00}&	\textbf{61.26}&	\textbf{35.50}&	\textbf{56.61}&	\textbf{40.72} &	\textbf{30.81}&	\textbf{46.78} \\
    \hline
    \hline
    Undistilled \tiny{\textit{Qwen3-VL-4B-Thinking}} & 43.93&	62.60&	49.37&	31.55&	49.86&	39.37&	32.08 & 44.11\\
    \cline{1-9}
    \end{tabular}
    }
    \vspace{-0.4cm}
    \label{tab:app_student_gen_2b}
\end{table}

\subsection{Computational Comparison with other VLM Distillations}
\label{app:subsec:computational}
\vspace{-0.2cm}
In Masking-KD, the auxiliary student forward pass increases computational overhead, and extracting attention maps incurs additional memory usage.
We investigate these aspects compared with other VLM distillation methods, including na"ive response distillation, LLaVA-KD~\cite{llava-kd}, CompoDistill~\cite{compodistill}, and Align-TI~\cite{align_kd}, in Tab.~\ref{tab:app_computational}.
For per-step time (s), we measure the time required to process 512 samples (one step in our training recipe) on two A100 GPUs, and report memory usage as the average over one step.
The reported average Pass@1 results come from Tab.~\ref{tab:main_vlm_distill_2b} of the main manuscript.
We use the 8B teacher and the 2B student from Qwen3-VL-Thinking.
Align-TI shows the highest computational overhead because it also requires extra forward passes and intermediate statistics.
Other VLM distillation methods incur only marginal additional memory usage because they mainly handle partial attention maps, such as visual attention maps.
In contrast, our method requires response-to-response attention maps over long reasoning traces, leading to relatively higher memory overhead.
Despite the increased overhead, Masking-KD achieves superior performance.

\begin{table}[h]
    \vspace{-0.2cm}
    \centering
    \scriptsize
    \renewcommand{\arraystretch}{1.2}
    \setlength{\tabcolsep}{6pt}   
        \caption{\textbf{Computational Comparison with other VLM distillations.} Per step (s) denotes the measured time required to process 512 samples (one step in our training recipe) on two A100 GPUs.}
        \vspace{0.1cm}
    \scalebox{1.15}{
    \begin{tabular}{l|cc||c}
    \hline
    Method     & Per step (s) & Memory (GB) & Avg. Pass@1  \\
    \hline
    Na\"ive Response Distillation     & 75.8  & 36.5 & 38.3  \\
    LLaVA-KD~\cite{llava-kd} & 100.4 &  37.2 & 40.3 \\
    CompoDistill~\cite{compodistill} & 103.9 & 38.0 & 41.5 \\
    Align-TI~\cite{align_kd} & \textbf{381.2} &   38.2 & 40.5 \\ \cdashline{1-4}[0.2pt/1pt]
    Masking-KD (ours) & 157.9 & \textbf{40.9} & \textbf{47.3} \\
    \hline
    \end{tabular}
    }
    \vspace{-0.3cm}
    \label{tab:app_computational}
\end{table}

\subsection{Statistical Significance}
\label{app:statistical}
\vspace{-0.2cm}
To ensure the statistical significance of our results, we additionally run Masking-KD three times with different random seeds.
Together with the original run reported in the main manuscript, we report the mean performance and standard deviation over four runs in total.
For this experiment, we use the 8B teacher and 2B student from Qwen3-VL-Thinking~\cite{qwen3_vl}.
As shown in Tab.~\ref{tab:app_statistical}, Masking-KD maintains low variance across runs.
These results indicate that the performance gains of our method are stable and not sensitive to random seeds.

\begin{table}[h!]
    \vspace{-0.2cm}
    \centering
    \scriptsize
    \renewcommand{\arraystretch}{1.2}
    \setlength{\tabcolsep}{4.1pt}   
    \vspace{-0.2cm}
        \caption{\textbf{Statistical Significance.}
    We additionally run Masking-KD three times and report the mean performance with standard deviation denoted by $\pm$ over four runs in total.
    }
    \vspace{0.1cm}
    \scalebox{1.055}{
    \begin{tabular}{m{1.7cm}|*{7}{>{\centering\arraybackslash}m{1.1cm}}|>{\centering\arraybackslash}m{1.1cm}}
    \cline{1-9}
    Trials     &  Geo3k & MathVista & We-Math & MMK12 & MathVerse & LogicVista & MMMU$^\text{Pro}$ & Avg. \\ \cline{1-9}
    Run 1 (reported) & 40.93	&59.20	&63.79&	37.20&	57.89	&41.61&	30.75 &	47.34 \\ \cdashline{1-9}[0.2pt/1pt]
    Run 2  & 43.93&	58.60&	63.56&	38.05&	57.98&	41.39&	30.81&	47.76 \\
    Run 3 &43.26	&58.00&	63.33&	36.45&	59.22&	41.39&	30.52&	47.45\\
    Run 4 &41.76	&58.70	&63.28&	36.85&	57.34&	42.06&	30.40&	47.20\\ \hline
    Avg. $\pm$std 
    & 42.47{\tiny{$\pm$1.16}} 
    & 58.63{\tiny{$\pm$0.50}} 
    & 63.49{\tiny{$\pm$0.20}} 
    & 37.14{\tiny{$\pm$0.71}} 
    & 58.11{\tiny{$\pm$0.71}} 
    & 41.61{\tiny{$\pm$0.27}} 
    & 30.62{\tiny{$\pm$0.18}} 
    & 47.44{\tiny{$\pm$0.23}} \\
    \cline{1-9}
    \end{tabular}
    }
    \vspace{-0.4cm}
    \label{tab:app_statistical}
\end{table}

\subsection{Evaluation on Hallucination Benchmark}
\label{app:subsec:pope}
\vspace{-0.2cm}
Since our method masks salient reasoning prefixes during distillation, one may be concerned that the masking could induce hallucination.
To examine this, we evaluate our method in a zero-shot manner on POPE~\cite{pope}, a widely used benchmark for object hallucination in VLMs.
For this experiment, we use the 8B teacher and 2B student from Qwen3-VL-Thinking~\cite{qwen3_vl}.
As shown in Tab.~\ref{tab:app_pope}, Masking-KD performs comparably to or slightly better than both the base model (\ie, the undistilled Qwen3-VL-2B-Thinking) and na\"ive distillation, indicating that our masking does not induce hallucination.

\begin{table}[h!]
    \vspace{-0.2cm}
    \centering
    \small
    \renewcommand{\arraystretch}{1.2}
    \setlength{\tabcolsep}{4.1pt}   
        \caption{\textbf{Zero-shot evaluation on hallucination benchmark.}
    We report accuracy (Acc) and F1 score on the random, popular, and adversarial splits of POPE~\cite{pope}.
    We use the 8B teacher and 2B student from Qwen3-VL-Thinking~\cite{qwen3_vl}.
    }
    \vspace{0.1cm}
    \scalebox{1.00}{
    \begin{tabular}{m{3.3cm}|*{2}{>{\centering\arraybackslash}m{0.9cm}}|*{2}{>{\centering\arraybackslash}m{0.9cm}}|*{2}{>{\centering\arraybackslash}m{0.9cm}}|*{2}{>{\centering\arraybackslash}m{0.9cm}}}
    \cline{1-9}
    \multirow{2}{*}{Method} & \multicolumn{2}{c|}{Random} & \multicolumn{2}{c|}{Popular} & \multicolumn{2}{c|}{Adversarial} & \multicolumn{2}{c}{Overall} \\
    & Acc & F1 & Acc & F1 & Acc & F1 & Acc & F1 \\ \cline{1-9}
    Base \tiny{\textit{{Qwen3-VL-2B-Thinking}}} & 90.27 & 89.41 & 88.17 & 87.43 & \textbf{87.17} & \textbf{86.60} & 88.53 & 87.80 \\
    Na\"ive Distillation & 90.83 & 90.13 & 88.50 & 87.91 & 86.87 & 86.41 & 88.73 & 88.12 \\ \cdashline{1-9}[0.2pt/1pt]
    Masking-KD (ours) & \textbf{91.17} & \textbf{90.57} & \textbf{88.53} & \textbf{88.08} & 86.60 & 86.37 & \textbf{88.77} & \textbf{88.31} \\
    \hline
    \end{tabular}
    }
    \vspace{-0.3cm}
    \label{tab:app_pope}
\end{table}

\section{Additional Ablation Studies}
\label{appendix:additional_ablation}
\vspace{-0.2cm}

\subsection{Ablation Study within Proposed Method.}
\label{app:subsec:ablation_within}
\vspace{-0.2cm}
Tab.~\ref{tab:ablation_within} ablates our key design choices and hyperparameters within the proposed methods.
For \textit{token-wise salient reasoning prefix masking} (Tab.~\ref{tab:ablation_within_a}), we observe:
\textbf{(1) token-wise adaptive masking}, where adaptive masking outperforms a non-adaptive masking, which masks the same prefixes at every decoding step;
and \textbf{(2) which reasoning prefixes to mask}, where masking high-attention prefixes achieves the best results over random, low-attention, and middle-attention masking.
For \textit{self-paced masking budget scheduling} (Tab.~\ref{tab:ablation_within_b}), we study:
\textbf{(3) threshold type}, where our cumulative ratio that masking prefixes according to their accumulated influence is more effective;
and \textbf{(4) the range of cummulative ratio}, where $\rho_{\text{min}}=0.3$ and $\rho_{\text{max}}=0.5$ yield the best performance.

\begin{table}[h!]
    \centering
    \footnotesize
    \renewcommand{\arraystretch}{1.2}
    \setlength{\tabcolsep}{8pt}
    \begin{minipage}{\linewidth}
    \centering
    \vspace{-0.3cm}
        \caption{\textbf{Ablation within each proposed method.}}
            \label{tab:ablation_within}
    \vspace{-0.1cm}
    \begin{subtable}[b]{0.48\linewidth}
        \centering
        \caption{Token-wise salient reasoning-prefix masking}
        \begin{minipage}[c]{\linewidth}
            \centering
                \scalebox{1.1}{
            \begin{tabular}{l|c}
               \hline
               Ablations  & Avg. \\
               \hline
               \multicolumn{2}{l}{~~(1) Effect of token-wise adaptive masking} \\
               \cdashline{1-2}[0.2pt/1pt]
               Non-adaptive masking & 41.73 \\ \cdashline{1-2}[0.2pt/1pt]
               Token-wise adaptive masking & \textbf{47.34} \\
               \hline
               \multicolumn{2}{l}{~~(2) Which reasoning prefixes to mask} \\
               \cdashline{1-2}[0.2pt/1pt]
               Random prefixes & 42.71 \\
               Low-attention prefixes& 37.29 \\ 
               Middle-attention prefixes& 38.84 \\ \cdashline{1-2}[0.2pt/1pt]
               High-attention (\ie, salient) prefixes & \textbf{47.34} \\
               \hline
            \end{tabular}
            }
        \end{minipage}
        \label{tab:ablation_within_a}
    \end{subtable}
    \hfill
    \begin{subtable}[b]{0.48\linewidth}
        \centering
        \caption{Self-paced masking budget scheduling}
        \begin{minipage}[t]{\linewidth}
            \centering
                \scalebox{1.1}{
            \begin{tabular}{l|c}
               \hline
               Ablations  & Avg. \\
               \hline
               \multicolumn{2}{l}{~~(3) Threshold type} \\
               \cdashline{1-2}[0.2pt/1pt]
               Attention threshold & 46.29 \\
               Masking ratio \% & 46.21 \\ \cdashline{1-2}[0.2pt/1pt]
               Cumulative ratio $\rho$ & \textbf{47.34} \\
               \hline
               \multicolumn{2}{l}{~~(4) $[\rho_{\text{min}},\rho_{\text{max}}]$ in cumulative ratio $\rho$} \\
               \cdashline{1-2}[0.2pt/1pt]
               {[0.1,0.3]} & 45.76 \\
               {[0.5,0.7]} & 45.82 \\ \cdashline{1-2}[0.2pt/1pt]
               {[0.3,0.5]} & \textbf{47.34} \\
               \hline
            \end{tabular}
            }
        \end{minipage}
        \label{tab:ablation_within_b}
    \end{subtable}

    \end{minipage}

    \vspace{-0.3cm}
\end{table}

\subsection{Auxiliary Student Forward}
\label{app:subsec:auxiliary}
\vspace{-0.2cm}
To construct the salient reasoning-prefix mask, we use an auxiliary student forward pass (Sec.~\ref{sec:auxiliary}) to obtain the response-to-response attention map $\mathbf{A}^{\text{reps}}$ and token-wise reverse KL divergence $\mathbf{r}$.
In Tab.~\ref{tab:app_auxiliary_weight_share}, we compare our weight-shared auxiliary student with a separately initialized frozen student.
The weight-shared design performs best, showing that adapting the mask construction to the current student state is effective.
\begin{table}[h!]
    \vspace{-0.2cm}
    \centering
    \scriptsize
    \renewcommand{\arraystretch}{1.2}
    \setlength{\tabcolsep}{4.1pt}   
    \caption{\textbf{Ablation within Auxiliary Student Forward}.
    $\dagger$ denotes using an auxiliary forward pass from a separately initialized frozen student, instead of the currently distilled student (reported).
    }
    \vspace{0.1cm}
    \scalebox{1.07}{
    \begin{tabular}{m{3cm}|*{7}{>{\centering\arraybackslash}m{0.94cm}}|>{\centering\arraybackslash}m{0.7cm}}
    \cline{1-9}
    Method     &  Geo3k & MathVista & We-Math & MMK12 & MathVerse & LogicVista & MMMU$^\text{Pro}$ & Avg. \\ \cline{1-9}
    No weight-sharing$^\dagger$  &  \textbf{43.93} &	58.40	&63.75	&34.20	&\textbf{58.85}	&40.04	&28.27&	46.78 \\ \cdashline{1-9}[0.2pt/1pt]
    Weight-sharing (reported) & 40.93 &	\textbf{59.20}	&\textbf{63.79}	&\textbf{37.20}&	57.89	&\textbf{41.61}	&\textbf{30.75}&	\textbf{47.34}\\
    \hline
    \end{tabular}
    }
    \vspace{-0.3cm}
    \label{tab:app_auxiliary_weight_share}
\end{table}

\subsection{Loss Functions}
\label{app:subsec:loss_functions}
\vspace{-0.2cm}
Our distillation framework is built upon reverse KL divergence~\cite{gkd} between the student's predictive distribution and the teacher's distribution, as described in Sec.~\ref{subsec:overall_framework}.
However, forward KL divergence and mixed KL objectives, which combine forward and reverse KL divergence with equal weights, are also widely used in distillation.
Here, we investigate the effect of these distillation losses in Tab.~\ref{tab:app_loss_functions}.
Using reverse KL divergence alone achieves the best results, indicating that its mode-seeking behavior~\cite{gkd} is more effective for encouraging the student to focus on the teacher’s high-probability predictions, rather than matching the entire distribution (\eg, forward KL div.).

\begin{table}[h!]
    \vspace{-0.2cm}
    \centering
    \scriptsize
    \renewcommand{\arraystretch}{1.2}
    \setlength{\tabcolsep}{4.1pt}   
        \caption{\textbf{Ablation on Loss Function}.
    $\dagger$ denotes the mixed KL objective, which combines forward KL divergence and reverse KL divergence with equal weights of 0.5.
    }
    \vspace{0.1cm}
    \scalebox{1.08}{
    \begin{tabular}{m{2.8cm}|*{7}{>{\centering\arraybackslash}m{0.94cm}}|>{\centering\arraybackslash}m{0.7cm}}
    \cline{1-9}
    Method     &  Geo3k & MathVista & We-Math & MMK12 & MathVerse & LogicVista & MMMU$^\text{Pro}$ & Avg. \\ \cline{1-9}
    Forward KL Div.  &  42.60&	57.00	&59.14	&35.35&	54.27&	38.70	&28.38	&45.06 \\
    Mixed KL Div.$^{\dagger}$ & \textbf{42.61}&	57.80&	60.92&	35.35&	56.65&	37.14&	30.17	&45.80\\  \cdashline{1-9}[0.2pt/1pt]
    Reverse KL Div. (reported)  &  40.93&	\textbf{59.20}	&\textbf{63.79}	&\textbf{37.20}&	\textbf{57.89}	&\textbf{41.61}	&\textbf{30.75}	&\textbf{47.34} \\
    \hline
    \end{tabular}
    }
    \vspace{-0.3cm}
    \label{tab:app_loss_functions}
\end{table}

\subsection{Excluding Immediate Previous Token from Masking}
\label{app:subsec:previous_token}
\vspace{-0.2cm}
As described in Implementation Details of Sec.~\ref{subsec:experiments_setup}, we exclude the immediate previous prefix token from masking to stabilize training and prevent loss explosion.
Here, we ablate this design in Tab.~\ref{tab:app_immediate_previous} by including the immediate previous prefix token in the masking candidates.
When this token is included in masking, the student struggles to predict subsequent tokens, significantly degrading performance.
Thus, excluding it from masking preserves essential local context for next-token prediction while maintaining training stability.

\begin{table}[h!]
    \vspace{-0.2cm}
    \centering
    \scriptsize
    \renewcommand{\arraystretch}{1.2}
    \setlength{\tabcolsep}{4.1pt}   
        \caption{\textbf{Ablation on excluding immediate previous token from masking.}
    As elaborated in Impplementation Details of Sec.~\ref{subsec:experiments_setup} from the main manuscript, we exclude this from masking to stable training and prevent the loss explosion, since excluding the immediate previous token from masking preserves essential local context for next-token prediction.
    }
    \vspace{0.1cm}
    \scalebox{1.07}{
    \begin{tabular}{m{3.1cm}|*{7}{>{\centering\arraybackslash}m{0.94cm}}|>{\centering\arraybackslash}m{0.7cm}}
    \cline{1-9}
    Method     &  Geo3k & MathVista & We-Math & MMK12 & MathVerse & LogicVista & MMMU$^\text{Pro}$ & Avg. \\ \cline{1-9}
    Including immediate previous token in masking &  \textbf{41.43} &	57.70	&54.89	&26.35&	51.06	&30.43&	24.51&	40.91 \\ \cdashline{1-9}[0.2pt/1pt]
    Excluding immediate previous token from masking (ours) & 40.93 &	\textbf{59.20}	&\textbf{63.79}	&\textbf{37.20}&	\textbf{57.89}	&\textbf{41.61}	&\textbf{30.75}&	\textbf{47.34}\\
    \hline
    \end{tabular}
    }
    \vspace{-0.3cm}
    \label{tab:app_immediate_previous}
\end{table}

\section{Additional Analyses}
\label{appendix:additional_analysis}
\vspace{-0.2cm}

\subsection{Effectiveness on Longer Reasoning Traces}
\label{app:subsec:trace_length}
\vspace{-0.2cm}
In this section, we investigate whether the advantage of Masking-KD becomes larger on longer reasoning traces by comparing traces of different total lengths through distillation with longer traces and length-binned analyses of performance and visual attention.
Since textual shortcut learning arises from reasoning prefixes accumulated over a trace (Sec.~\ref{subsec:effect_masking}), its effect, and thus the benefit of Masking-KD, is expected to grow with trace length.
Note that the analyses in Fig.~\ref{fig:visual_forgetting_warp} and Fig.~\ref{fig:reverse_kl_div} of the main manuscript examine changes along relative positions within each trace, rather than across traces of different total lengths.
For all experiments, we use the 8B teacher and 2B student from Qwen3-VL-Thinking~\cite{qwen3_vl}.

\paragraph{Distillation with Longer Traces.}
We first study whether Masking-KD benefits further from longer reasoning traces. To this end, we construct the distillation data from teacher traces generated with a maximum length of 8,192 tokens, in addition to the default setting with a maximum length of 4,096 tokens (Sec.~\ref{subsec:experiments_setup}).
This yields 24K correct traces, from which we randomly sample 19K traces to match the size of the default training set for a fair comparison.
As shown in Tab.~\ref{tab:app_long_trace}, Masking-KD obtains a clear gain when trained on longer traces (47.34 $\rightarrow$ 51.01 in average Pass@1), indicating that the advantage of our approach is further amplified on longer reasoning traces.

\begin{table}[h!]
    \vspace{-0.2cm}
    \centering
    \scriptsize
    \renewcommand{\arraystretch}{1.2}
    \setlength{\tabcolsep}{4.1pt}   
        \caption{\textbf{Distillation with longer reasoning traces.}
    We construct the distillation data from teacher traces generated with a maximum length of 8,192 tokens, instead of 4,096 tokens (default), and randomly sample 19K correct traces to match the size of the default training set.
    }
    \vspace{0.1cm}
    \scalebox{1.055}{
    \begin{tabular}{m{1.7cm}|*{7}{>{\centering\arraybackslash}m{1.1cm}}|>{\centering\arraybackslash}m{1.1cm}}
    \cline{1-9}
    Max. Length     &  Geo3k & MathVista & We-Math & MMK12 & MathVerse & LogicVista & MMMU$^\text{Pro}$ & Avg. \\ \cline{1-9}
    4,096 (default) & 40.93 & \textbf{59.20} & 63.79 & 37.20 & 57.89 & 41.61 & 30.75 & 47.34 \\ \cdashline{1-9}[0.2pt/1pt]
    8,192 & \textbf{46.09} & 58.50 & \textbf{66.38} & \textbf{46.45} & \textbf{64.17} & \textbf{41.83} & \textbf{33.64} & \textbf{51.01} \\
    \hline
    \end{tabular}
    }
    \vspace{-0.3cm}
    \label{tab:app_long_trace}
\end{table}

\paragraph{Trace-Length Distribution.}
In Tab.~\ref{tab:app_trace_length_dist}, we report the trace-length distribution of the training data to clarify how trace lengths differ between the two settings and whether the default training set is affected by truncation.
The default training set contains no truncated traces, since a trace that reaches the 4,096-token limit fails to produce a final answer and is therefore removed when we keep only correct responses (Sec.~\ref{subsec:experiments_setup}).
Thus, all 19K training traces in the default setting contain complete final answers within the 4,096-token limit, while the 8,192-token setting provides substantially longer traces (\eg, 4,013 vs.~7,818 tokens at p99).

\begin{table}[h!]
    \vspace{-0.2cm}
    \centering
    \small
    \renewcommand{\arraystretch}{1.2}
    \setlength{\tabcolsep}{4.1pt}   
        \caption{\textbf{Trace-length distribution of training data.}
    We report the percentiles of the number of tokens in the training traces for each setting.
    For the 8,192-token setting, the statistics are computed over the 19K traces used for training.
    }
    \vspace{0.1cm}
    \scalebox{1.00}{
    \begin{tabular}{m{2.5cm}|*{5}{>{\centering\arraybackslash}m{1.1cm}}}
    \cline{1-6}
    Max. Length     &  p1 & p25 & p50 & p75 & p99 \\ \cline{1-6}
    4,096 (default) & 221 & 951 & 1,552 & 2,492 & 4,013 \\ \cdashline{1-6}[0.2pt/1pt]
    8,192 & 236 & 1,096 & 1,978 & 3,648 & 7,818 \\
    \hline
    \end{tabular}
    }
    \vspace{-0.3cm}
    \label{tab:app_trace_length_dist}
\end{table}

\paragraph{Performance by Trace Length.}
We further analyze the relationship between trace length and performance gains by comparing the average Pass@1 of na\"ive distillation and Masking-KD across trace-length bins in Tab.~\ref{tab:app_trace_length_bin}.
For a paired comparison, we assign each evaluation example to a bin according to the length of the response generated by the 8B teacher, so that both methods are evaluated on exactly the same examples within each bin.
Across the first three bins, the gain of Masking-KD gradually increases as the trace length grows (+7.55, +20.43, and +27.77).
The last bin (3,072+) is strongly affected by the 4,096-token evaluation limit, as many generations in this bin reach the maximum length without producing a final answer and are therefore counted as incorrect.
We thus consider the result in this bin inconclusive under the current evaluation limit, rather than evidence against the length-dependent trend.

\paragraph{Visual Attention by Trace Length.}
In Tab.~\ref{tab:app_trace_length_bin}, we also compare the average visual attention of na\"ive distillation and Masking-KD using the same bins and evaluation examples, verifying whether the student's use of visual evidence also depends on trace length.
While both methods show comparable visual attention in the shortest-trace bin, Masking-KD achieves consistently higher visual attention across all longer-trace bins (+12.9\% to +17.2\%), further supporting its advantage on longer reasoning traces.

\begin{table}[h!]
    \vspace{-0.2cm}
    \centering
    \small
    \renewcommand{\arraystretch}{1.2}
    \setlength{\tabcolsep}{4.1pt}   
        \caption{\textbf{Average Pass@1 and visual attention by trace length.}
    Each evaluation example is assigned to a bin according to the length of the response generated by the 8B teacher, so that na\"ive distillation and Masking-KD are evaluated on exactly the same examples within each bin.
    Gain denotes the difference between Masking-KD and na\"ive distillation, and relative gain denotes this difference normalized by na\"ive distillation.
    }
    \vspace{0.1cm}
    \scalebox{1.00}{
    \begin{tabular}{m{2.6cm}|*{4}{>{\centering\arraybackslash}m{1.9cm}}}
    \cline{1-5}
    Trace Length     &  0--1,024 & 1,024--2,048 & 2,048--3,072 & 3,072+ \\ \cline{1-5}
    \multicolumn{5}{l}{\textit{Average Pass@1}} \\ \cdashline{1-5}[0.2pt/1pt]
    Na\"ive Distillation & 69.85 & 40.82 & 18.32 & \textbf{8.89} \\
    Masking-KD (ours) & \textbf{77.40} & \textbf{61.25} & \textbf{46.09} & 8.43 \\
    Gain & +7.55 & +20.43 & +27.77 & $-$0.46 \\
    \cline{1-5}
    \multicolumn{5}{l}{\textit{Average Visual Attention}} \\ \cdashline{1-5}[0.2pt/1pt]
    Na\"ive Distillation & \textbf{0.0309} & 0.0207 & 0.0157 & 0.0140 \\
    Masking-KD (ours) & 0.0300 & \textbf{0.0242} & \textbf{0.0184} & \textbf{0.0158} \\
    Relative Gain & $-$2.9\% & +16.9\% & +17.2\% & +12.9\% \\
    \hline
    \end{tabular}
    }
    \vspace{-0.3cm}
    \label{tab:app_trace_length_bin}
\end{table}

\subsection{Analysis on Self-Paced Masking Budget}
\label{app:subsec:rho_analysis}
\vspace{-0.2cm}
In this section, we analyze how our self-paced masking budget scheduling (Sec.~\ref{subsec:self-paced}) allocates the masking budget $\rho_n$ and how it differs from a static budget.
For comparison, we consider a static-budget setting with a fixed $\rho_n=0.4$ for all tokens, which is the midpoint of our range $[\rho_{\text{min}},\rho_{\text{max}}]=[0.3,0.5]$.
For all experiments, we use the 8B teacher and 2B student from Qwen3-VL-Thinking~\cite{qwen3_vl}.

\paragraph{Distillation Signal over Training.}
We first compare the distillation signal of the two settings over training progress in Tab.~\ref{tab:app_rho_signal}, where the signal is measured by the masked-context reverse KL divergence (\ie, $\mathcal{L}_{\text{Distill}}$ in Eq.~\eqref{eq:reverse_kl}).
In both settings, the signal naturally decays as optimization proceeds.
However, self-paced scheduling consistently maintains a stronger signal throughout training, leading to higher average Pass@1 (45.91 $\rightarrow$ 47.34).

\begin{table}[h!]
    \vspace{-0.2cm}
    \centering
    \small
    \renewcommand{\arraystretch}{1.2}
    \setlength{\tabcolsep}{4.1pt}   
        \caption{\textbf{Distillation signal over training.}
    We report the masked-context reverse KL divergence (\ie, $\mathcal{L}_{\text{Distill}}$ in Eq.~\eqref{eq:reverse_kl}) averaged over each interval of training progress.
    The static budget uses a fixed $\rho_n=0.4$, the midpoint of $[\rho_{\text{min}},\rho_{\text{max}}]$.
    }
    \vspace{0.1cm}
    \scalebox{1.00}{
    \begin{tabular}{m{2.7cm}|*{4}{>{\centering\arraybackslash}m{1.3cm}}|>{\centering\arraybackslash}m{1.6cm}}
    \cline{1-6}
    \multirow{2}{*}{Masking Budget} & \multicolumn{4}{c|}{Training Progress} & \multirow{2}{*}{Avg. Pass@1} \\
    & 0--25\% & 25--50\% & 50--75\% & 75--100\% & \\ \cline{1-6}
    Static ($\rho_n=0.4$) & 20.41 & 13.56 & 9.46 & 8.44 & 45.91 \\ \cdashline{1-6}[0.2pt/1pt]
    Self-paced (ours) & \textbf{21.47} & \textbf{15.10} & \textbf{10.70} & \textbf{9.60} & \textbf{47.34} \\
    \hline
    \end{tabular}
    }
    \vspace{-0.3cm}
    \label{tab:app_rho_signal}
\end{table}

\paragraph{Mask Overlap between Static and Self-Paced Budgets.}
To identify where this stronger signal comes from, we analyze the overlap of the masked prefixes between the two settings.
The two settings mask a nearly identical total amount of prefixes, with a ratio of 0.998 between the total masking amounts of the self-paced and static budgets.
However, the masked prefixes differ substantially, with a Jaccard similarity of 0.700 between the two masked token sets.
This indicates that the benefit of self-paced scheduling comes from masking different prefixes, rather than masking more prefixes.

\paragraph{Budget Allocation across Decoding Positions.}
In Tab.~\ref{tab:app_rho_position}, we report how the masking budget is allocated across decoding positions by measuring the average $\rho_n$ across prefix-length bins.
As the prefix grows, the full-context reverse KL divergence $r_n$ decreases, indicating that later tokens become easier to imitate from the accumulated context.
Accordingly, unlike the static budget, our self-paced scheduling gradually increases $\rho_n$ along the trace (0.385 $\rightarrow$ 0.429), applying stronger masking to later positions where imitation becomes easier and the distillation signal weakens.
As a result, the masked-context reverse KL divergence also increases along the trace (20.79 $\rightarrow$ 30.40), consistent with stronger perturbation at later decoding positions.

\begin{table}[h!]
    \vspace{-0.2cm}
    \centering
    \small
    \renewcommand{\arraystretch}{1.2}
    \setlength{\tabcolsep}{4.1pt}   
        \caption{\textbf{Budget allocation across decoding positions.}
    Each response token is assigned to a bin according to its prefix length (\ie, the number of preceding response tokens), and statistics are averaged over 10K training traces.
    The full-context reverse KL divergence corresponds to $r_n$ in Eq.~\eqref{eq:r}, which is used for scheduling, and the masked-context reverse KL divergence corresponds to $\mathcal{L}_{\text{Distill}}$ in Eq.~\eqref{eq:reverse_kl}.
    }
    \vspace{0.1cm}
    \scalebox{1.00}{
    \begin{tabular}{m{4.4cm}|*{4}{>{\centering\arraybackslash}m{1.7cm}}}
    \cline{1-5}
    Prefix Length     &  0--1,024 & 1,024--2,048 & 2,048--3,072 & 3,072+ \\ \cline{1-5}
    Full-context Reverse KL Div. & 1.015 & 0.833 & 0.715 & 0.450 \\ \cdashline{1-5}[0.2pt/1pt]
    $\rho_n$ (Static) & 0.400 & 0.400 & 0.400 & 0.400 \\
    $\rho_n$ (Self-paced, ours) & 0.385 & 0.394 & 0.405 & 0.429 \\ \cdashline{1-5}[0.2pt/1pt]
    Masked-context Reverse KL Div. & 20.79 & 23.65 & 25.11 & 30.40 \\
    \hline
    \end{tabular}
    }
    \vspace{-0.3cm}
    \label{tab:app_rho_position}
\end{table}

\paragraph{Budget Allocation across Token Types.}
To verify whether the masking budget adapts to the availability of visual evidence, we classify each predicted token into function words, verbs, and content words using the spaCy part-of-speech tagger~\cite{spacy}, and report per-type statistics in Tab.~\ref{tab:app_rho_token_type}.
Function words and verbs show lower visual attention, whereas content words rely more on the image.
Accordingly, our self-paced scheduling applies weaker masking to function words and verbs ($\rho_n=0.363$ and $0.357$) and stronger masking to content words ($\rho_n=0.410$).
Moreover, the masked-context reverse KL divergence is also higher for content words, showing that masking concentrates its learning signal on tokens where visual evidence matters.

\begin{table}[h!]
    \vspace{-0.2cm}
    \centering
    \small
    \renewcommand{\arraystretch}{1.2}
    \setlength{\tabcolsep}{4.1pt}   
        \caption{\textbf{Budget allocation across predicted token types.}
    Each predicted token is classified by the spaCy part-of-speech tagger~\cite{spacy}, and statistics are averaged over the 19K training traces.
    Visual attention denotes the average visual attention when predicting each token type.
    }
    \vspace{0.1cm}
    \scalebox{1.00}{
    \begin{tabular}{m{4.4cm}|*{3}{>{\centering\arraybackslash}m{2.0cm}}}
    \cline{1-4}
    Token Type     &  Function Words & Verbs & Content Words \\ \cline{1-4}
    Visual Attention & 0.0144 & 0.0134 & 0.0152 \\ \cdashline{1-4}[0.2pt/1pt]
    $\rho_n$ (Static) & 0.400 & 0.400 & 0.400 \\
    $\rho_n$ (Self-paced, ours) & 0.363 & 0.357 & 0.410 \\ \cdashline{1-4}[0.2pt/1pt]
    Masked-context Reverse KL Div. & 18.65 & 18.55 & 23.51 \\
    \hline
    \end{tabular}
    }
    \vspace{-0.3cm}
    \label{tab:app_rho_token_type}
\end{table}

\newpage
\begin{wrapfigure}{r}{0.4\textwidth}
    \centering
    \vspace{-0.4cm}
    \includegraphics[width=\linewidth]{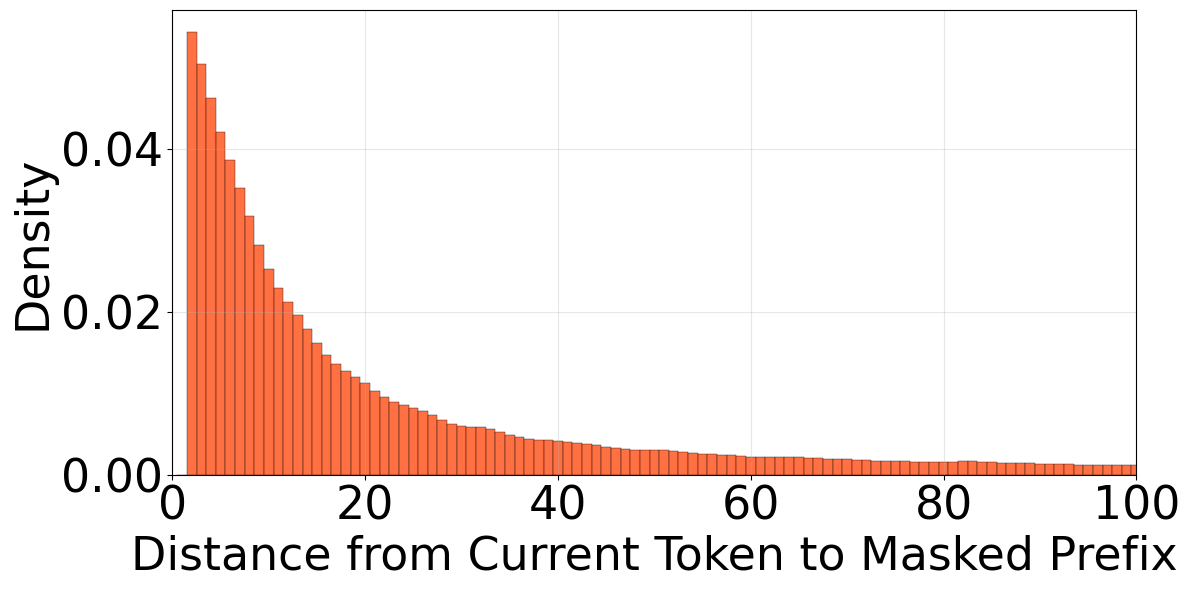}
    \vspace{-0.53cm}
    \caption{Masked prefix distance}
    \vspace{-0.4cm}
    \label{fig:app_masked_position}
\end{wrapfigure}
\subsection{Statistics of Masked Prefix Positions}
\label{app:subsec:masked_position}
\vspace{-0.2cm}
In this section, we analyze the relative position of masked prefixes with respect to the current token (\ie, the distance from the current token to the masked prefix) over 19k teacher responses, as shown in Fig.~\ref{fig:app_masked_position}.
As described in the Implementation Details of Sec.~\ref{subsec:experiments_setup} and Appendix~\ref{app:subsec:previous_token}, we exclude the immediate previous prefix token from masking (\eg, distance zero) to stabilize training.
We observe that masking is more frequently applied to prefixes closer to the current token, resulting in a higher density of masked tokens in the recent context.
As the distance from the current token increases, the masking frequency gradually decreases.
This suggests that highly influential (\ie, salient) reasoning prefixes are often located near the current token, and our masking strategy effectively targets these contexts.



\vspace{-0.2cm}
\subsection{Example of Inference}
\label{app:subsec:inference_example}
\begin{figure}[h!] 
    \centering
    \vspace{-0.1cm}
    \includegraphics[width=1\linewidth]{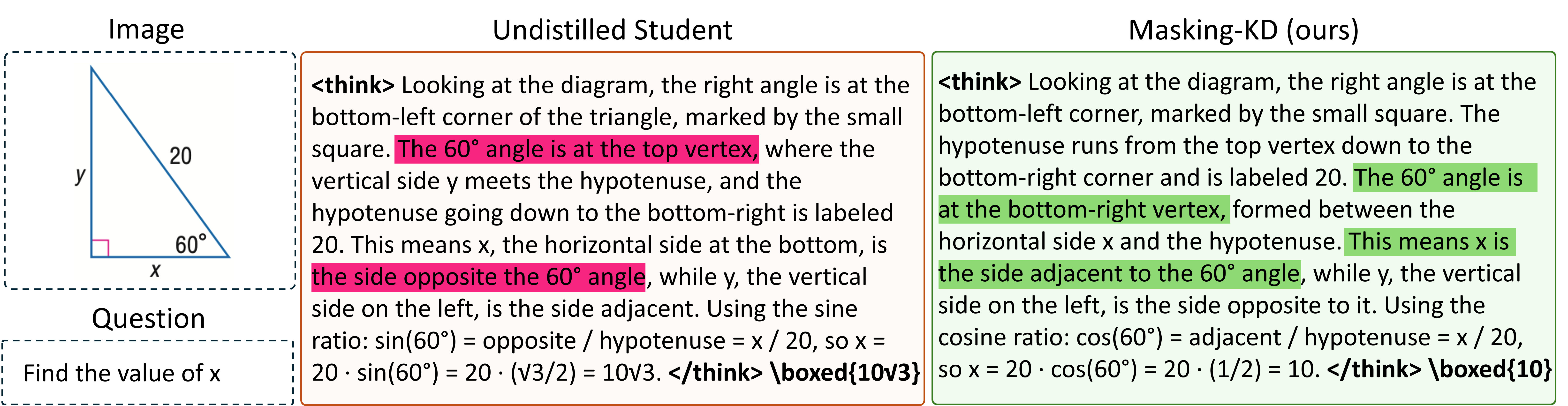}
    \caption{\textbf{Example of the Inference.} 
    }
    \label{fig:app_qual_inference}
\end{figure}

\vspace{-0.2cm}
Fig.~\ref{fig:app_qual_inference} illustrates the think-answer response of our Masking-KD compared with the undistilled student (\ie, Qwen3-VL-2B-Thinking).
Undistilled student produce perception errors, highlighted in \colorbox{red!40}{red box}, whereas ours show enhanced visual perception, highlighted in \colorbox{green!40}{green box}.

\subsection{More Comparison on Visual Attention Map}
\label{app:subsec:visual_attn_comp}
\begin{figure}[h!] 
    \centering
    \includegraphics[width=1\linewidth]{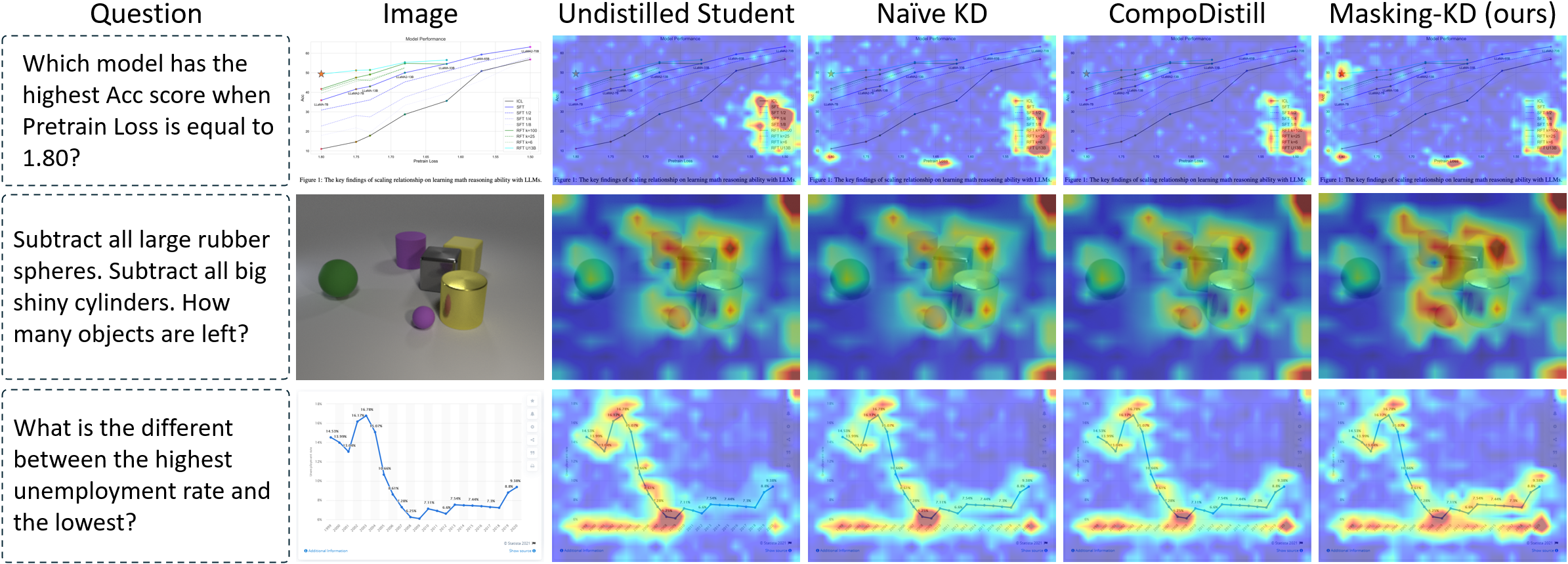}
    \caption{\textbf{More Comparison on Visual Attention Map.}
    We average the visual attention scores over the entire thinking trace.}
    \label{fig:app_overall_visual_attn}
\end{figure}

\vspace{-0.2cm}
We provide additional comparisons on the visual attention map in Fig.~\ref{fig:app_overall_visual_attn}.
Compared with all compared methods, Masking-KD produces more focused attention on semantically relevant image regions, indicating that the student relies more on visual evidence throughout the reasoning process.

\subsection{More Prediction Behavior of the Student during Distillation.}
\label{app:subsec:predict_behavior}
\begin{figure}[h!]
\captionsetup[subfigure]{skip=-2pt}
    \centering
    \begin{subfigure}{\linewidth}
        \centering
        \includegraphics[width=\linewidth]{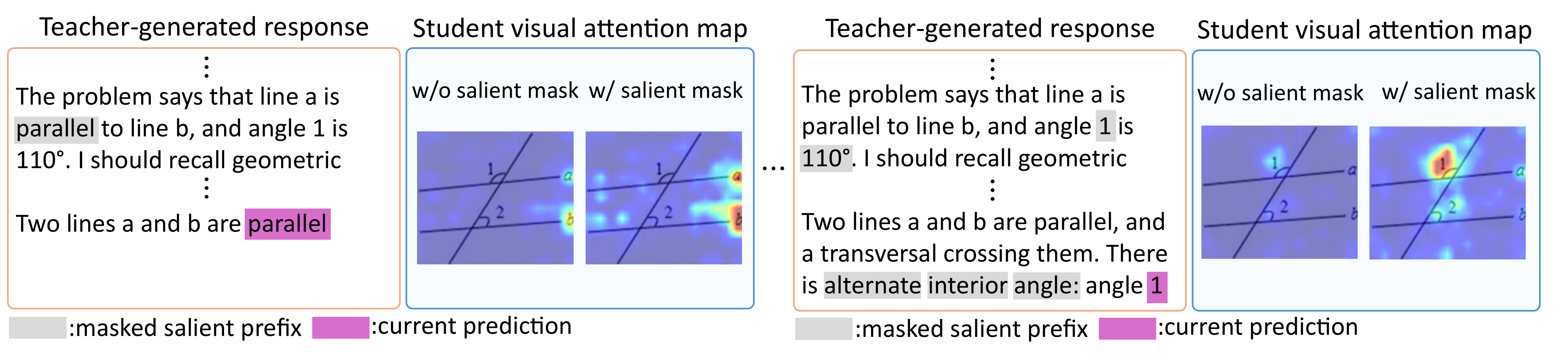}
        \caption{math}
        \label{fig:qual_math}
    \end{subfigure}

    \begin{subfigure}{\linewidth}
        \centering
        \includegraphics[width=\linewidth]{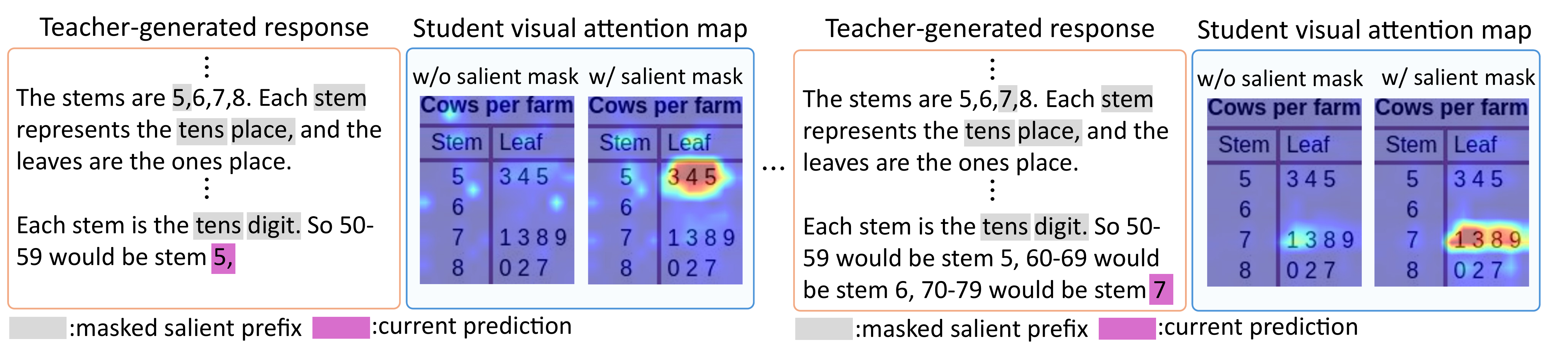}
        \caption{STEM}
        \label{fig:qual_stem}
    \end{subfigure}

    \begin{subfigure}{\linewidth}
        \centering
        \includegraphics[width=\linewidth]{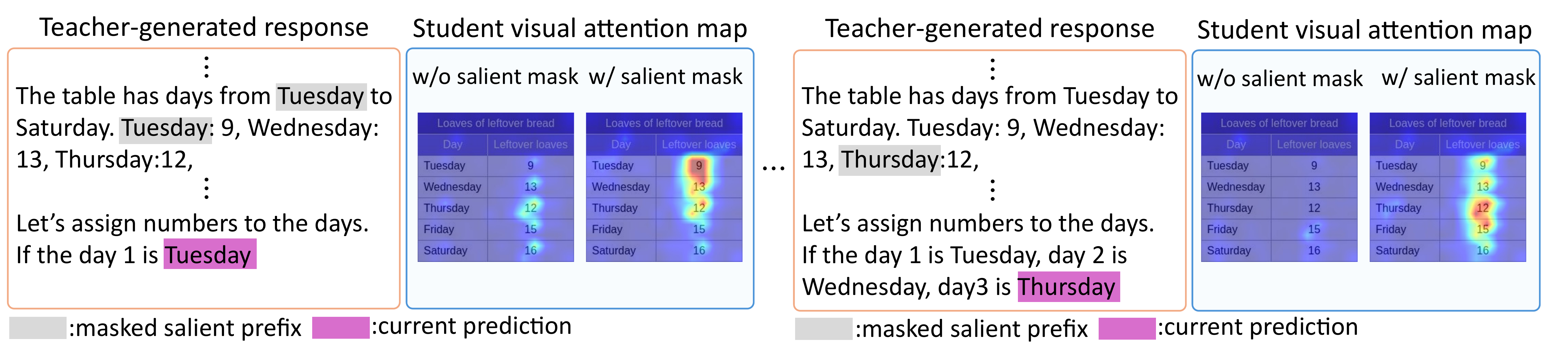}
        \caption{table}
        \label{fig:qual_table}
    \end{subfigure}
    
    \begin{subfigure}{\linewidth}
        \centering
        \includegraphics[width=\linewidth]{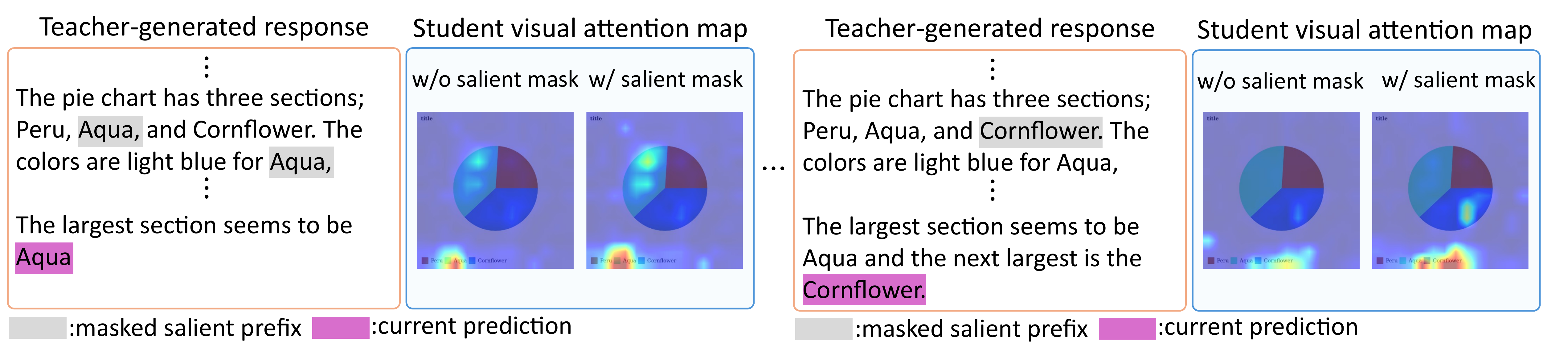}
        \caption{chart}
        \label{fig:qual_chart}
    \end{subfigure}
    
    \vspace{-0.1cm}
    \caption{\textbf{More Prediction Behavior of the Student during Distillation} without and with our salient reasoning-prefix mask across four types of reasoning problems: (a) math, (b) STEM, (c) table, and (d) chart.}
    \vspace{-0.3cm}
    \label{fig:app_qual}
\end{figure}

\vspace{-0.2cm}
We illustrate the prediction behavior of the student during distillation without and with our salient reasoning-prefix mask in Fig.~\ref{fig:qual_distill} of the main manuscript.
In this section, we provide additional qualitative results across four types of reasoning problems: (a) math, (b) STEM, (c) table and (d) chart in Fig.~\ref{fig:app_qual}.
These examples further show that our masking strategy encourages the student to attend to relevant visual regions when predicting the current token, rather than relying solely on exposed textual reasoning prefixes.


\section{Additional Details}
\label{appendix:implementation_details}
\vspace{-0.2cm}

\subsection{Details on Top-$\rho$ Masking}
\label{app:subsec:top-rho}
\vspace{-0.2cm}
To select salient prefixes to mask, we use a nucleus top-$p$ style rule~\cite{top_p} (\ie, top-$\rho_n$ masking), as described in Eq.~\eqref{eq:cumulative} of the main manuscript and rewritten below:
\begin{equation}
    \label{app:eq:cumulative}
    {\sum_{j\in\mathcal{S}_n}\mathbf{\bar{A}}^{\text{resp}}_{n,j}} \geq \rho_n, \quad \text{where}~\bar{\mathbf{A}}^{\text{resp}}_{n,j}=\frac{\mathbf{A}^{\text{resp}}_{n,j}}{\sum_{k=1}^{n-1}\mathbf{A}^{\text{resp}}_{n,k}}.
\end{equation}
In this section, we detail this top-$\rho_n$ masking step-by-step.

For each $n$-th row in response-to-response attention map ${\mathbf{A}}^{\text{resp}}$, we first normalize its attention over preceding response tokens and sort the prefix tokens in descending order $\downarrow$ of attention, as follows:
\begin{equation}
    \pi_n=\mathrm{argsort}_{j<n}^{\downarrow}(\bar{\mathbf{A}}^{\text{resp}})_{n,j},\quad \text{where~}\bar{\mathbf{A}}^{\text{resp}}_{n,j}=\frac{\mathbf{A}^{\text{resp}}_{n,j}}{\sum_{k=1}^{n-1}\mathbf{A}^{\text{resp}}_{n,k}}.
\end{equation}
We then collect the top-ranked prefix tokens until their cumulative attention mass reaches the self-paced cumulative ratio $\rho_n$:
\begin{equation}
    \label{app:eq:cumulative_2}
    \mathcal{S}_n=\{\pi_n(1), \dots, \pi_n(k_n)\}, \quad \text{where~} k_n=\min \Bigg\{K\Bigg|\sum_{i=1}^K {\bar{\mathbf{A}}}^{\text{resp}}_{n,\pi_n(i)} \geq \rho_n \Bigg\}.
\end{equation}
Here, $\bar{\mathbf{A}}^{\text{resp}}_{n,\pi_n(i)}$ denotes the attention assigned by response token $y_n$ to the $i$-th highest-ranked prefix token under the ordering $\pi_n$.
The resulting salient prefix set $\mathcal{S}_n$ for all $n=\{1, \dots, N\}$ is then used to construct the salient reasoning-prefix mask $\tilde{\mathbf{M}}$ in Eq.~\eqref{eq:mask}.

\subsection{Details on Self-Distillation}
\label{app:subsec:self_distill}
\vspace{-0.2cm}
Our Masking-KD can operate under self-distillation settings, where the student serves as its own teacher.
In self-distillation, the full-context student prediction is detached and used as the teacher target, while the masked-context student prediction is optimized to match it.
This can be written by modifying the distillation loss in Eq.~\eqref{eq:reverse_kl} as:
\begin{equation}
\label{eq:app_reverse_kl}
\mathcal{L}_{\text{Distill}}
= \frac{1}{N}\sum_{n=1}^{N} \sum_{y\in\mathcal{V}} p_s({y} \mid \mathbf{x}_v,\mathbf{x}_q,\mathbf{y}_{<n},\tilde{\mathbf{M}})\log\frac{p_s({y} \mid \mathbf{x}_v,\mathbf{x}_q,\mathbf{y}_{<n},\tilde{\mathbf{M}})}{\text{StopGrad}\big(p_s({y} \mid \mathbf{x}_v,\mathbf{x}_q,\mathbf{y}_{<n},\mathbf{M})\big)}.
\end{equation}
Here, \text{StopGrad($\cdot$)} denotes the stop-gradient operation, so the full-context branch serves as a fixed self-teacher target during optimization.
All other hyperparameters and training recipes follow the knowledge distillation setting described in the Implementation Details of Sec.~\ref{subsec:experiments_setup}.

\subsection{Instruction for Teacher-generated Response}
\label{app:instruction}
\begin{figure}[h!] 
    \centering
    \includegraphics[width=0.8\linewidth]{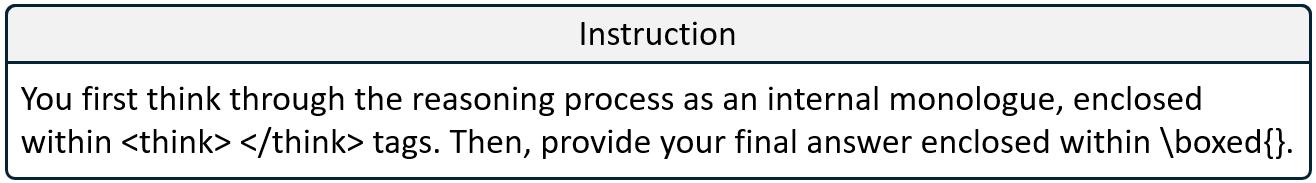}
    \caption{The instruction is used to prompt the teacher model to generate think-answer trajectories for distillation data.}
    \label{fig:app_instruction}
\end{figure}

\vspace{-0.2cm}
To extract the teacher's think-answer trajectories from ViRL39k~\cite{rethinker} dataset, we use the instruction illustrated in Fig.~\ref{fig:app_instruction}.
The instruction is appended after the image and question to prompt the teacher model to generate think-answer trajectories.

\section{Further Discussion}
\label{appendix:further_discussion}
\vspace{-0.2cm}

\subsection{Limitation}
\label{appendix:limitation}
\vspace{-0.2cm}
While Masking-KD introduces additional computational overhead and memory usage due to the auxiliary student forward pass and response-to-response attention map extraction, the overhead remains manageable in practice, as reported in Tab.~\ref{tab:app_computational} of Appendix~\ref{app:subsec:computational}.
In this work, we mainly focus on improving visually anchored reasoning during distillation rather than optimizing training efficiency.
Further reducing the computational cost of salient prefix selection remains an important direction for future work.

\subsection{Social Impact}
\label{appendix:social_impact}
\vspace{-0.2cm}
Our work improves the efficiency of think-answer VLMs by transferring reasoning capabilities to compact student models, which may help reduce deployment costs and broaden accessibility.
However, like other VLMs, the distilled models may still inherit biases or generate incorrect outputs.
Careful evaluation and responsible deployment are therefore important when applying these models in real-world settings.



\clearpage

\section*{NeurIPS Paper Checklist}


\begin{enumerate}

\item {\bf Claims}
    \item[] Question: Do the main claims made in the abstract and introduction accurately reflect the paper's contributions and scope?
    \item[] Answer: \answerYes{} 
    \item[] Justification: The contributions summarized in Sec.~\ref{sec:intro} are detailed.
    \item[] Guidelines:
    \begin{itemize}
        \item The answer \answerNA{} means that the abstract and introduction do not include the claims made in the paper.
        \item The abstract and/or introduction should clearly state the claims made, including the contributions made in the paper and important assumptions and limitations. A \answerNo{} or \answerNA{} answer to this question will not be perceived well by the reviewers. 
        \item The claims made should match theoretical and experimental results, and reflect how much the results can be expected to generalize to other settings. 
        \item It is fine to include aspirational goals as motivation as long as it is clear that these goals are not attained by the paper. 
    \end{itemize}

\item {\bf Limitations}
    \item[] Question: Does the paper discuss the limitations of the work performed by the authors?
    \item[] Answer: \answerYes{} 
    \item[] Justification: We discuss the limitations of this work in Appendix~\ref{appendix:limitation}.
    \item[] Guidelines:
    \begin{itemize}
        \item The answer \answerNA{} means that the paper has no limitation while the answer \answerNo{} means that the paper has limitations, but those are not discussed in the paper. 
        \item The authors are encouraged to create a separate ``Limitations'' section in their paper.
        \item The paper should point out any strong assumptions and how robust the results are to violations of these assumptions (e.g., independence assumptions, noiseless settings, model well-specification, asymptotic approximations only holding locally). The authors should reflect on how these assumptions might be violated in practice and what the implications would be.
        \item The authors should reflect on the scope of the claims made, e.g., if the approach was only tested on a few datasets or with a few runs. In general, empirical results often depend on implicit assumptions, which should be articulated.
        \item The authors should reflect on the factors that influence the performance of the approach. For example, a facial recognition algorithm may perform poorly when image resolution is low or images are taken in low lighting. Or a speech-to-text system might not be used reliably to provide closed captions for online lectures because it fails to handle technical jargon.
        \item The authors should discuss the computational efficiency of the proposed algorithms and how they scale with dataset size.
        \item If applicable, the authors should discuss possible limitations of their approach to address problems of privacy and fairness.
        \item While the authors might fear that complete honesty about limitations might be used by reviewers as grounds for rejection, a worse outcome might be that reviewers discover limitations that aren't acknowledged in the paper. The authors should use their best judgment and recognize that individual actions in favor of transparency play an important role in developing norms that preserve the integrity of the community. Reviewers will be specifically instructed to not penalize honesty concerning limitations.
    \end{itemize}

\item {\bf Theory assumptions and proofs}
    \item[] Question: For each theoretical result, does the paper provide the full set of assumptions and a complete (and correct) proof?
    \item[] Answer: \answerNA{} 
    \item[] Justification: We do not present theoretical results or formal proofs in this paper.
    \item[] Guidelines:
    \begin{itemize}
        \item The answer \answerNA{} means that the paper does not include theoretical results. 
        \item All the theorems, formulas, and proofs in the paper should be numbered and cross-referenced.
        \item All assumptions should be clearly stated or referenced in the statement of any theorems.
        \item The proofs can either appear in the main paper or the supplemental material, but if they appear in the supplemental material, the authors are encouraged to provide a short proof sketch to provide intuition. 
        \item Inversely, any informal proof provided in the core of the paper should be complemented by formal proofs provided in appendix or supplemental material.
        \item Theorems and Lemmas that the proof relies upon should be properly referenced. 
    \end{itemize}

    \item {\bf Experimental result reproducibility}
    \item[] Question: Does the paper fully disclose all the information needed to reproduce the main experimental results of the paper to the extent that it affects the main claims and/or conclusions of the paper (regardless of whether the code and data are provided or not)?
    \item[] Answer: \answerYes{} 
    \item[] Justification: We explain experimental setups in Sec.~\ref{subsec:experiments_setup}.
    \item[] Guidelines:
    \begin{itemize}
        \item The answer \answerNA{} means that the paper does not include experiments.
        \item If the paper includes experiments, a \answerNo{} answer to this question will not be perceived well by the reviewers: Making the paper reproducible is important, regardless of whether the code and data are provided or not.
        \item If the contribution is a dataset and\slash or model, the authors should describe the steps taken to make their results reproducible or verifiable. 
        \item Depending on the contribution, reproducibility can be accomplished in various ways. For example, if the contribution is a novel architecture, describing the architecture fully might suffice, or if the contribution is a specific model and empirical evaluation, it may be necessary to either make it possible for others to replicate the model with the same dataset, or provide access to the model. In general. releasing code and data is often one good way to accomplish this, but reproducibility can also be provided via detailed instructions for how to replicate the results, access to a hosted model (e.g., in the case of a large language model), releasing of a model checkpoint, or other means that are appropriate to the research performed.
        \item While NeurIPS does not require releasing code, the conference does require all submissions to provide some reasonable avenue for reproducibility, which may depend on the nature of the contribution. For example
        \begin{enumerate}
            \item If the contribution is primarily a new algorithm, the paper should make it clear how to reproduce that algorithm.
            \item If the contribution is primarily a new model architecture, the paper should describe the architecture clearly and fully.
            \item If the contribution is a new model (e.g., a large language model), then there should either be a way to access this model for reproducing the results or a way to reproduce the model (e.g., with an open-source dataset or instructions for how to construct the dataset).
            \item We recognize that reproducibility may be tricky in some cases, in which case authors are welcome to describe the particular way they provide for reproducibility. In the case of closed-source models, it may be that access to the model is limited in some way (e.g., to registered users), but it should be possible for other researchers to have some path to reproducing or verifying the results.
        \end{enumerate}
    \end{itemize}

\item {\bf Open access to data and code}
    \item[] Question: Does the paper provide open access to the data and code, with sufficient instructions to faithfully reproduce the main experimental results, as described in supplemental material?
    \item[] Answer: \answerYes{} 
    \item[] Justification: Codes are included in the supplementary material Zip file. We use datasets that are publicly available. We will release our code on GitHub in the future.
    \item[] Guidelines:
    \begin{itemize}
        \item The answer \answerNA{} means that paper does not include experiments requiring code.
        \item Please see the NeurIPS code and data submission guidelines (\url{https://neurips.cc/public/guides/CodeSubmissionPolicy}) for more details.
        \item While we encourage the release of code and data, we understand that this might not be possible, so \answerNo{} is an acceptable answer. Papers cannot be rejected simply for not including code, unless this is central to the contribution (e.g., for a new open-source benchmark).
        \item The instructions should contain the exact command and environment needed to run to reproduce the results. See the NeurIPS code and data submission guidelines (\url{https://neurips.cc/public/guides/CodeSubmissionPolicy}) for more details.
        \item The authors should provide instructions on data access and preparation, including how to access the raw data, preprocessed data, intermediate data, and generated data, etc.
        \item The authors should provide scripts to reproduce all experimental results for the new proposed method and baselines. If only a subset of experiments are reproducible, they should state which ones are omitted from the script and why.
        \item At submission time, to preserve anonymity, the authors should release anonymized versions (if applicable).
        \item Providing as much information as possible in supplemental material (appended to the paper) is recommended, but including URLs to data and code is permitted.
    \end{itemize}

\item {\bf Experimental setting/details}
    \item[] Question: Does the paper specify all the training and test details (e.g., data splits, hyperparameters, how they were chosen, type of optimizer) necessary to understand the results?
    \item[] Answer: \answerYes{} 
    \item[] Justification: We provide all necessary details to understand the results in Sec.~\ref{sec:experiments}.
    \item[] Guidelines:
    \begin{itemize}
        \item The answer \answerNA{} means that the paper does not include experiments.
        \item The experimental setting should be presented in the core of the paper to a level of detail that is necessary to appreciate the results and make sense of them.
        \item The full details can be provided either with the code, in appendix, or as supplemental material.
    \end{itemize}

\item {\bf Experiment statistical significance}
    \item[] Question: Does the paper report error bars suitably and correctly defined or other appropriate information about the statistical significance of the experiments?
    \item[] Answer: \answerYes{} 
    \item[] Justification: We report statistical significance of the experiments in Appendix~\ref{app:statistical}.
    \item[] Guidelines:
    \begin{itemize}
        \item The answer \answerNA{} means that the paper does not include experiments.
        \item The authors should answer \answerYes{} if the results are accompanied by error bars, confidence intervals, or statistical significance tests, at least for the experiments that support the main claims of the paper.
        \item The factors of variability that the error bars are capturing should be clearly stated (for example, train/test split, initialization, random drawing of some parameter, or overall run with given experimental conditions).
        \item The method for calculating the error bars should be explained (closed form formula, call to a library function, bootstrap, etc.)
        \item The assumptions made should be given (e.g., Normally distributed errors).
        \item It should be clear whether the error bar is the standard deviation or the standard error of the mean.
        \item It is OK to report 1-sigma error bars, but one should state it. The authors should preferably report a 2-sigma error bar than state that they have a 96\% CI, if the hypothesis of Normality of errors is not verified.
        \item For asymmetric distributions, the authors should be careful not to show in tables or figures symmetric error bars that would yield results that are out of range (e.g., negative error rates).
        \item If error bars are reported in tables or plots, the authors should explain in the text how they were calculated and reference the corresponding figures or tables in the text.
    \end{itemize}

\item {\bf Experiments compute resources}
    \item[] Question: For each experiment, does the paper provide sufficient information on the computer resources (type of compute workers, memory, time of execution) needed to reproduce the experiments?
    \item[] Answer: \answerYes{}  
    \item[] Justification: We include the information on the computer resources in Sec.~\ref{subsec:experiments_setup}, Table~\ref{tab:app_computational}.
    \item[] Guidelines:
    \begin{itemize}
        \item The answer \answerNA{} means that the paper does not include experiments.
        \item The paper should indicate the type of compute workers CPU or GPU, internal cluster, or cloud provider, including relevant memory and storage.
        \item The paper should provide the amount of compute required for each of the individual experimental runs as well as estimate the total compute. 
        \item The paper should disclose whether the full research project required more compute than the experiments reported in the paper (e.g., preliminary or failed experiments that didn't make it into the paper). 
    \end{itemize}
    
\item {\bf Code of ethics}
    \item[] Question: Does the research conducted in the paper conform, in every respect, with the NeurIPS Code of Ethics \url{https://neurips.cc/public/EthicsGuidelines}?
    \item[] Answer:  \answerYes{} 
    \item[] Justification: This work conducted with the NeurIPS Code of Ethics.
    \item[] Guidelines:
    \begin{itemize}
        \item The answer \answerNA{} means that the authors have not reviewed the NeurIPS Code of Ethics.
        \item If the authors answer \answerNo, they should explain the special circumstances that require a deviation from the Code of Ethics.
        \item The authors should make sure to preserve anonymity (e.g., if there is a special consideration due to laws or regulations in their jurisdiction).
    \end{itemize}

\item {\bf Broader impacts}
    \item[] Question: Does the paper discuss both potential positive societal impacts and negative societal impacts of the work performed?
    \item[] Answer: \answerYes{} 
    \item[] Justification: We discuss social impacts in Sec.~\ref{appendix:social_impact}.
    \item[] Guidelines:
    \begin{itemize}
        \item The answer \answerNA{} means that there is no societal impact of the work performed.
        \item If the authors answer \answerNA{} or \answerNo, they should explain why their work has no societal impact or why the paper does not address societal impact.
        \item Examples of negative societal impacts include potential malicious or unintended uses (e.g., disinformation, generating fake profiles, surveillance), fairness considerations (e.g., deployment of technologies that could make decisions that unfairly impact specific groups), privacy considerations, and security considerations.
        \item The conference expects that many papers will be foundational research and not tied to particular applications, let alone deployments. However, if there is a direct path to any negative applications, the authors should point it out. For example, it is legitimate to point out that an improvement in the quality of generative models could be used to generate Deepfakes for disinformation. On the other hand, it is not needed to point out that a generic algorithm for optimizing neural networks could enable people to train models that generate Deepfakes faster.
        \item The authors should consider possible harms that could arise when the technology is being used as intended and functioning correctly, harms that could arise when the technology is being used as intended but gives incorrect results, and harms following from (intentional or unintentional) misuse of the technology.
        \item If there are negative societal impacts, the authors could also discuss possible mitigation strategies (e.g., gated release of models, providing defenses in addition to attacks, mechanisms for monitoring misuse, mechanisms to monitor how a system learns from feedback over time, improving the efficiency and accessibility of ML).
    \end{itemize}
    
\item {\bf Safeguards}
    \item[] Question: Does the paper describe safeguards that have been put in place for responsible release of data or models that have a high risk for misuse (e.g., pre-trained language models, image generators, or scraped datasets)?
    \item[] Answer: \answerNA{} 
    \item[] Justification: This paper does not pose such risks.
    \item[] Guidelines:
    \begin{itemize}
        \item The answer \answerNA{} means that the paper poses no such risks.
        \item Released models that have a high risk for misuse or dual-use should be released with necessary safeguards to allow for controlled use of the model, for example by requiring that users adhere to usage guidelines or restrictions to access the model or implementing safety filters. 
        \item Datasets that have been scraped from the Internet could pose safety risks. The authors should describe how they avoided releasing unsafe images.
        \item We recognize that providing effective safeguards is challenging, and many papers do not require this, but we encourage authors to take this into account and make a best faith effort.
    \end{itemize}

\item {\bf Licenses for existing assets}
    \item[] Question: Are the creators or original owners of assets (e.g., code, data, models), used in the paper, properly credited and are the license and terms of use explicitly mentioned and properly respected?
    \item[] Answer: \answerYes{} 
    \item[] Justification: We properly cite all assets and mention the license and terms of usage.
    \item[] Guidelines:
    \begin{itemize}
        \item The answer \answerNA{} means that the paper does not use existing assets.
        \item The authors should cite the original paper that produced the code package or dataset.
        \item The authors should state which version of the asset is used and, if possible, include a URL.
        \item The name of the license (e.g., CC-BY 4.0) should be included for each asset.
        \item For scraped data from a particular source (e.g., website), the copyright and terms of service of that source should be provided.
        \item If assets are released, the license, copyright information, and terms of use in the package should be provided. For popular datasets, \url{paperswithcode.com/datasets} has curated licenses for some datasets. Their licensing guide can help determine the license of a dataset.
        \item For existing datasets that are re-packaged, both the original license and the license of the derived asset (if it has changed) should be provided.
        \item If this information is not available online, the authors are encouraged to reach out to the asset's creators.
    \end{itemize}

\item {\bf New assets}
    \item[] Question: Are new assets introduced in the paper well documented and is the documentation provided alongside the assets?
    \item[] Answer: \answerYes{} 
    \item[] Justification: We are releasing our code as a new asset, fully documented on GitHub, to complement the documentation in this paper.
    \item[] Guidelines:
    \begin{itemize}
        \item The answer \answerNA{} means that the paper does not release new assets.
        \item Researchers should communicate the details of the dataset\slash code\slash model as part of their submissions via structured templates. This includes details about training, license, limitations, etc. 
        \item The paper should discuss whether and how consent was obtained from people whose asset is used.
        \item At submission time, remember to anonymize your assets (if applicable). You can either create an anonymized URL or include an anonymized zip file.
    \end{itemize}

\item {\bf Crowdsourcing and research with human subjects}
    \item[] Question: For crowdsourcing experiments and research with human subjects, does the paper include the full text of instructions given to participants and screenshots, if applicable, as well as details about compensation (if any)? 
    \item[] Answer: \answerNA{} 
    \item[] Justification: The paper does not involve crowdsourcing nor research with human subjects.
    \item[] Guidelines:
    \begin{itemize}
        \item The answer \answerNA{} means that the paper does not involve crowdsourcing nor research with human subjects.
        \item Including this information in the supplemental material is fine, but if the main contribution of the paper involves human subjects, then as much detail as possible should be included in the main paper. 
        \item According to the NeurIPS Code of Ethics, workers involved in data collection, curation, or other labor should be paid at least the minimum wage in the country of the data collector. 
    \end{itemize}

\item {\bf Institutional review board (IRB) approvals or equivalent for research with human subjects}
    \item[] Question: Does the paper describe potential risks incurred by study participants, whether such risks were disclosed to the subjects, and whether Institutional Review Board (IRB) approvals (or an equivalent approval/review based on the requirements of your country or institution) were obtained?
    \item[] Answer: \answerNA{} 
    \item[] Justification: This work does not involve human subjects, and therefore IRB approval is not required.
    \item[] Guidelines:
    \begin{itemize}
        \item The answer \answerNA{} means that the paper does not involve crowdsourcing nor research with human subjects.
        \item Depending on the country in which research is conducted, IRB approval (or equivalent) may be required for any human subjects research. If you obtained IRB approval, you should clearly state this in the paper. 
        \item We recognize that the procedures for this may vary significantly between institutions and locations, and we expect authors to adhere to the NeurIPS Code of Ethics and the guidelines for their institution. 
        \item For initial submissions, do not include any information that would break anonymity (if applicable), such as the institution conducting the review.
    \end{itemize}

\item {\bf Declaration of LLM usage}
    \item[] Question: Does the paper describe the usage of LLMs if it is an important, original, or non-standard component of the core methods in this research? Note that if the LLM is used only for writing, editing, or formatting purposes and does \emph{not} impact the core methodology, scientific rigor, or originality of the research, declaration is not required.
    \item[] Answer: \answerNA{}. 
    \item[] Justification: LMs/VLMs are used as the subject of study rather than as a tool for developing the method.
    \item[] Guidelines:
    \begin{itemize}
        \item The answer \answerNA{} means that the core method development in this research does not involve LLMs as any important, original, or non-standard components.
        \item Please refer to our LLM policy in the NeurIPS handbook for what should or should not be described.
    \end{itemize}

\end{enumerate}

\end{document}